\documentclass[journal]{vgtc}

\usepackage[utf8]{inputenc}
\PassOptionsToPackage{hyphens}{url}\usepackage{hyperref}
\usepackage{graphicx}
\usepackage{microtype}                 
\usepackage{textcomp}                  
\usepackage{times}                     
\usepackage{mathptmx}                  
\usepackage{cite}                      
\usepackage{bm}                        
\usepackage{amsmath}                   
\usepackage{amssymb}                   
\usepackage{amsthm}                    
\usepackage{url}
\usepackage{soul}

\usepackage[caption=false,font=sf]{subfig}
\usepackage{color}
\usepackage{xcolor}
\usepackage{tikz}
\usetikzlibrary{positioning}
\usepackage{empheq}
\usepackage[linesnumbered, ruled]{algorithm2e}

\usepackage{adjustbox}
\usepackage{array}
\newcolumntype{R}[2]{%
    >{\adjustbox{angle=#1,lap=\width-(#2)}\bgroup}%
    l%
    <{\egroup}%
}

\newcommand{\plotframe}[1]{{%
  \definecolor{borderColor}{rgb}{0.85, 0.85, 0.85}%
  \definecolor{backgroundColor}{rgb}{1,1,1}%
  \setlength{\fboxsep}{0.5pt}%
  \setlength{\fboxrule}{0.7pt}%
  \fcolorbox{borderColor}{backgroundColor}{#1}}%
}

\newcommand{\eqframe}[2]{
  \begin{empheq}[box={\fboxsep=12pt\fbox}]{align}\label{#1}#2\end{empheq}
}

\preprinttext{To appear in IEEE Transactions on Visualization and Computer Graphics.}
\onlineid{1124}

\vgtccategory{Research}
\vgtcpapertype{algorithm/technique}

\title{Uncertainty-Aware Principal Component Analysis}
\author{Jochen Görtler, Thilo Spinner, Dirk Streeb, Daniel Weiskopf, and Oliver Deussen}
\authorfooter{
 \item J. Görtler, T. Spinner, D. Streeb, and O. Deussen\\
       are with the University of Konstanz, Germany.\\
       E-mail: \{firstname.lastname\}@uni-konstanz.de
 \item D. Weiskopf is with the University of Stuttgart, Germany.\\
       E-mail: weiskopf@visus.uni-stuttgart.de
}

\newcommand{\Real}{\mathbb{R}}

\newcommand{\Span}[1]{\langle{#1}\rangle}

\renewcommand{\Vec}[1]{\vec{#1}}
\newcommand{\Mat}[1]{#1}
\newcommand{\Rvec}[1]{\mathbf{#1}}

\newcommand{\Zero}{\Vec{0}}

\newcommand{\Transpose}[1]{{#1}^{\mathsf{T}}}
\DeclareMathOperator{\Diag}{diag}
\DeclareMathOperator{\Ln}{ln}
\DeclareMathOperator{\Dim}{dim}

\DeclareMathOperator{\CovOp}{Cov}
\newcommand{\Cov}[2]{\CovOp(#1,#2)}
\newcommand{\Mean}[1]{\mu_{#1}}

\newcommand{\OuterP}[2]{{#1}\Transpose{#2}}
\newcommand{\OuterS}[1]{\OuterP{#1}{#1}}

\newcommand{\Normal}{N}
\newcommand{\DistMean}{\boldsymbol{\mu}}
\newcommand{\DistCov}{\boldsymbol{\Psi}}
\newcommand{\ExpCov}{\Mat{K}_{\ObSet\ObSet}}

\newcommand{\EvOper}{\mathbb{E}}
\newcommand{\Ev}[1]{\EvOper\left[#1\right]}
\newcommand{\EvCov}[1]{\hat{\EvOper}\left[#1\right]}
\newcommand{\EvOb}[1]{\Ev{#1}}

\newcommand{\Ob}{\Rvec{t}}
\newcommand{\ObSet}{\Rvec{T}}
\newcommand{\Error}{\Rvec{\boldsymbol{\epsilon}}}
\newcommand{\Follows}{\sim}

\newcommand{\Centering}[1]{\Mean{#1}\Transpose{\Mean{#1}}}

\newcommand{\Eigenvalue}{\lambda}
\newcommand{\Eigenvector}{\Vec{v}}
\newcommand{\PrincipalComp}{\mathbf{w}}
\newcommand{\PrincipalCompSet}{\mathbf{W}}
\newcommand{\ProjOper}{\Phi}
\newcommand{\Proj}[1]{\ProjOper(#1)}
\newcommand{\ProjColumnMat}{\mathbf{A}}

\newcommand{\BiLin}[2]{\Transpose{\Vec{#1}}\Mat{#2}\Vec{#1}}

\newcommand{\BigO}[1]{O(#1)}

\newcommand{\todo}[2][]{\texttt{\color[rgb]{1.0, 0.0, 0.0}{[todo\ifx&#1&\else@#1\fi:#2]}}}

\abstract{%
We present a technique to perform dimensionality reduction on data that is subject to uncertainty.
Our method is a generalization of traditional principal component analysis (PCA) to multivariate probability distributions.
In comparison to non-linear methods, linear dimensionality reduction techniques have the advantage that the characteristics of such probability distributions remain intact after projection.
We derive a representation of the PCA sample covariance matrix that respects potential uncertainty in each of the inputs, building the mathematical foundation of our new method: \emph{uncertainty-aware PCA}.
In addition to the accuracy and performance gained by our approach over sampling-based strategies, our formulation allows us to perform sensitivity analysis with regard to the uncertainty in the data.
For this, we propose \emph{factor traces} as a novel visualization that enables to better understand the influence of uncertainty on the chosen principal components.
We provide multiple examples of our technique using real-world datasets.
As a special case, we show how to propagate multivariate normal distributions through PCA in closed form.
Furthermore, we discuss extensions and limitations of our approach.
}


\teaser{
   \centering
  \newcommand{\ProgFigA}{\plotframe{\includegraphics[width=0.23\linewidth]{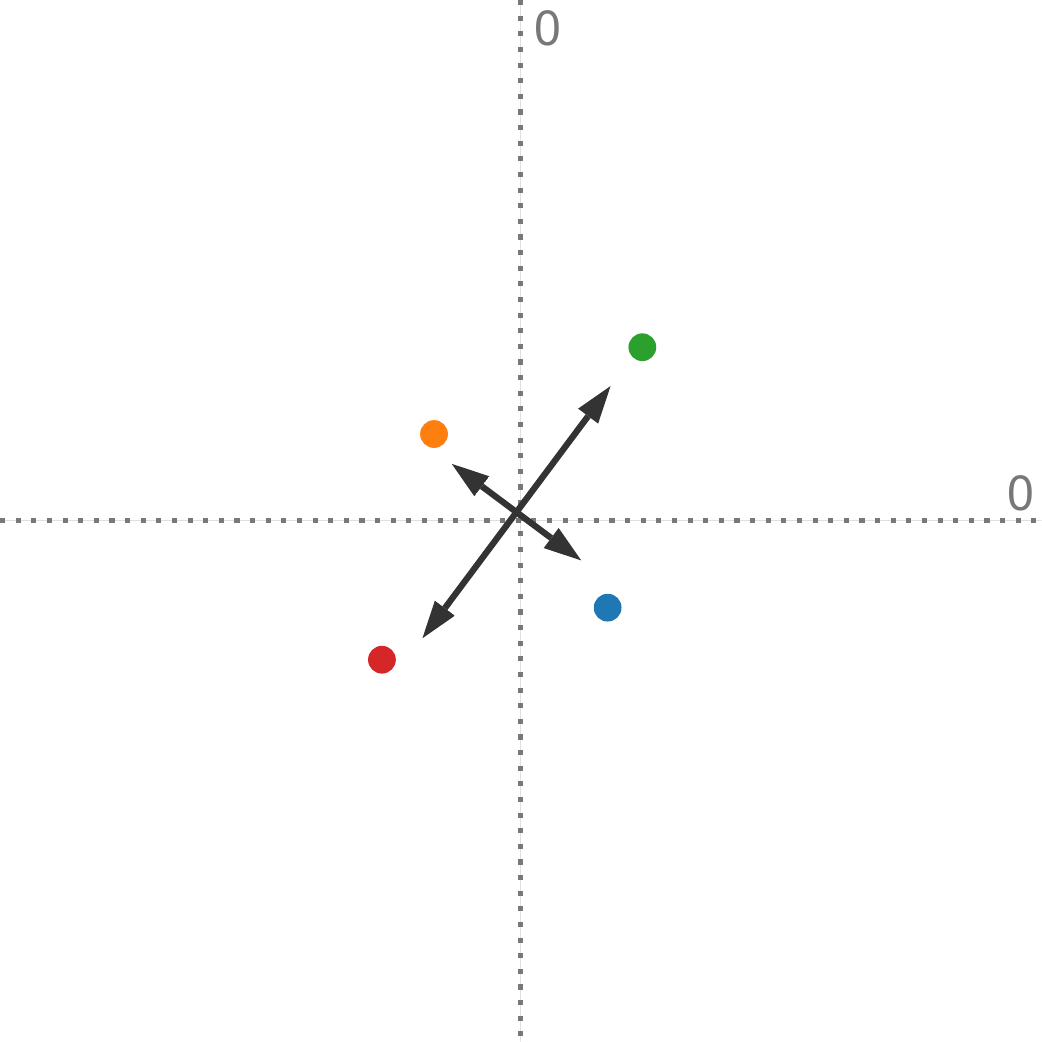}}}
\newcommand{\ProgFigB}{\plotframe{\includegraphics[width=0.23\linewidth]{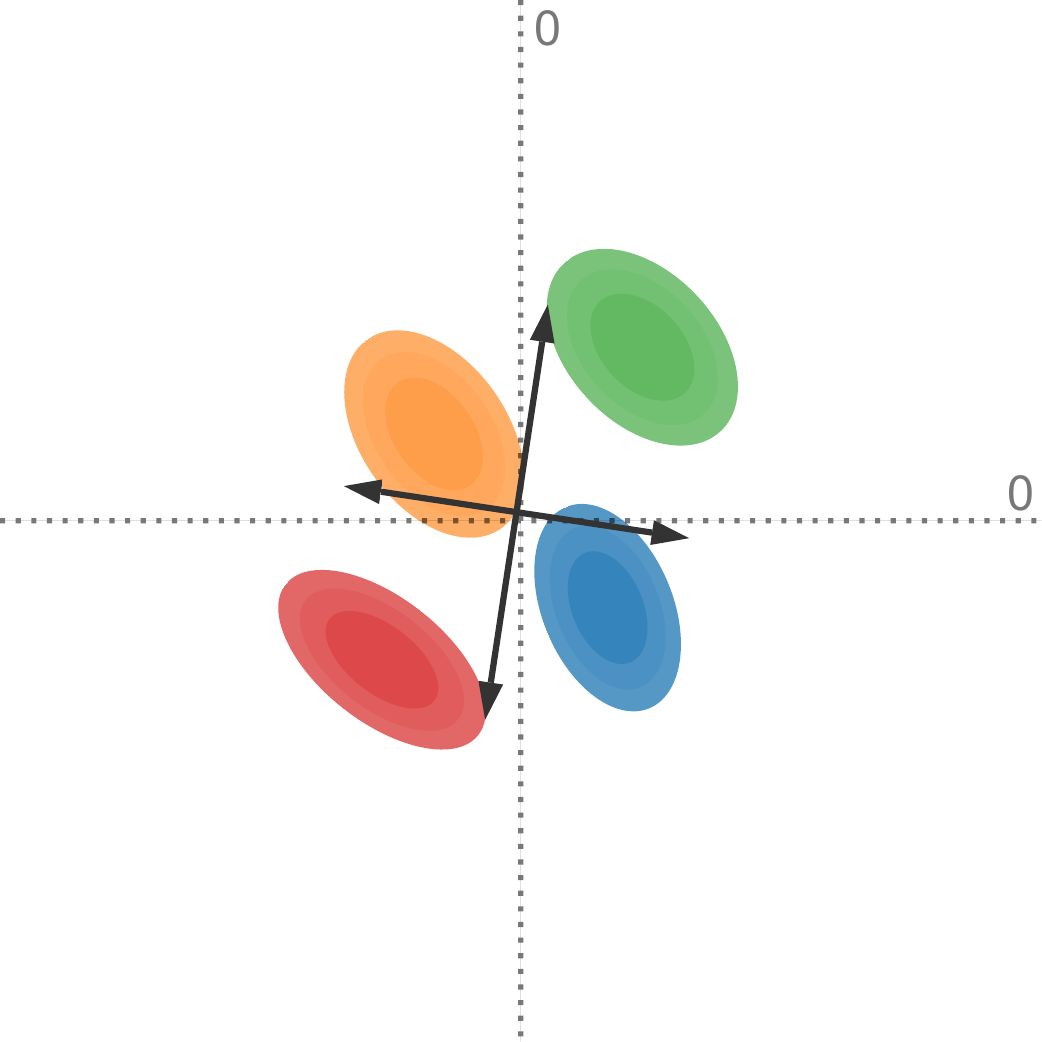}}}
\newcommand{\ProgFigC}{\plotframe{\includegraphics[width=0.23\linewidth]{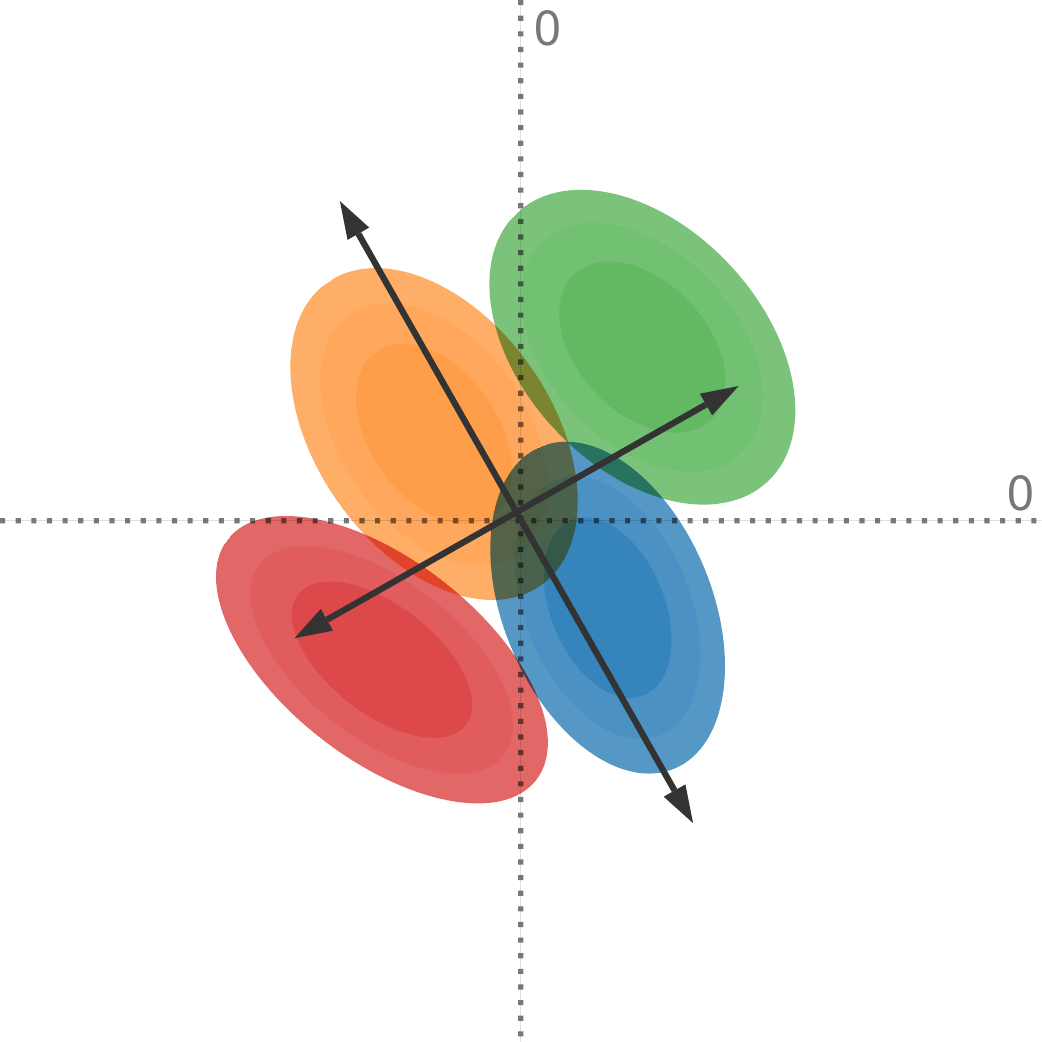}}}
\newcommand{\ProgFigD}{\plotframe{\includegraphics[width=0.23\linewidth]{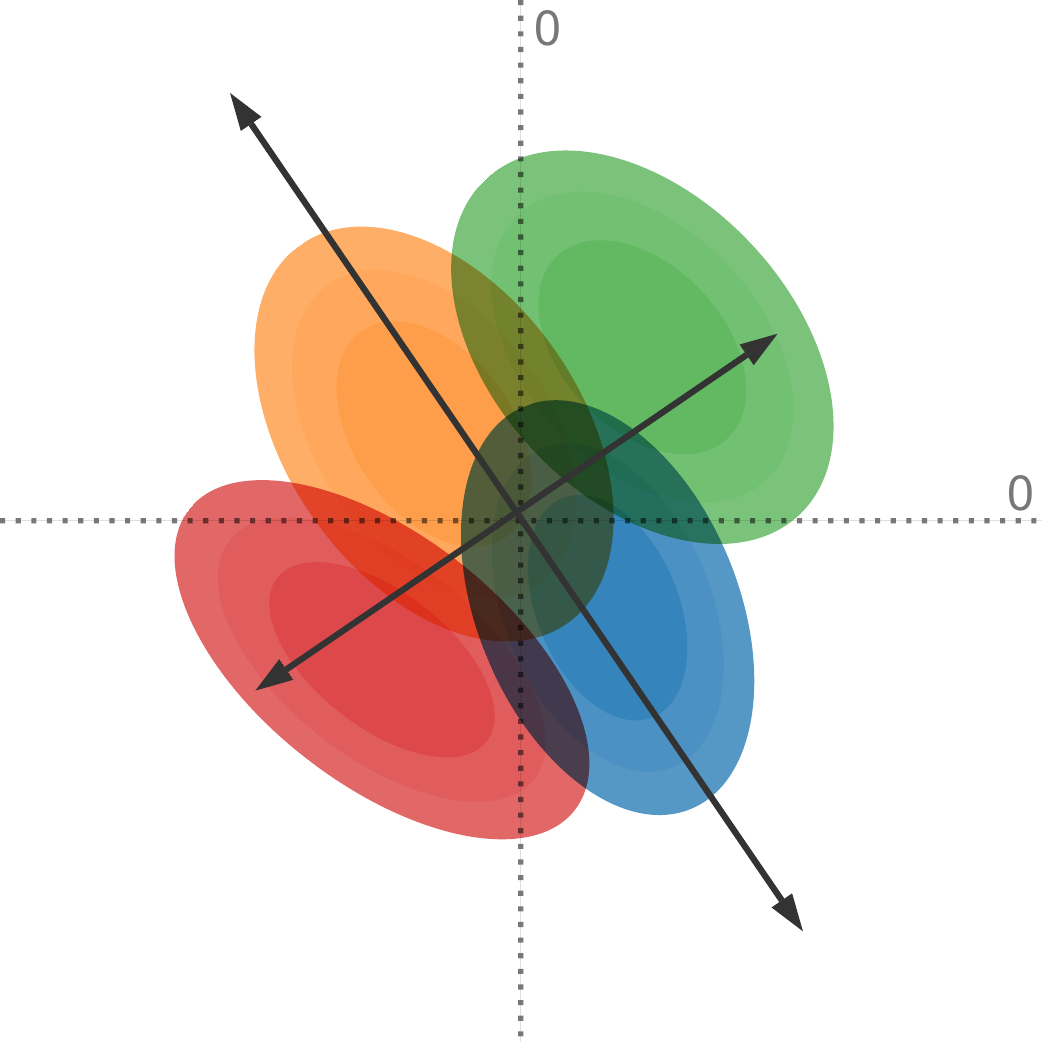}}}

\newcommand{\ProgFigAp}{\plotframe{\includegraphics[width=0.23\linewidth]{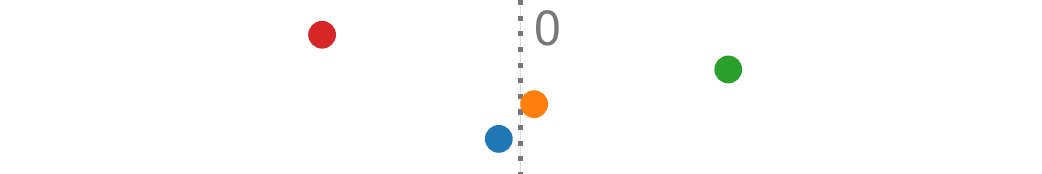}}}
\newcommand{\ProgFigBp}{\plotframe{\includegraphics[width=0.23\linewidth]{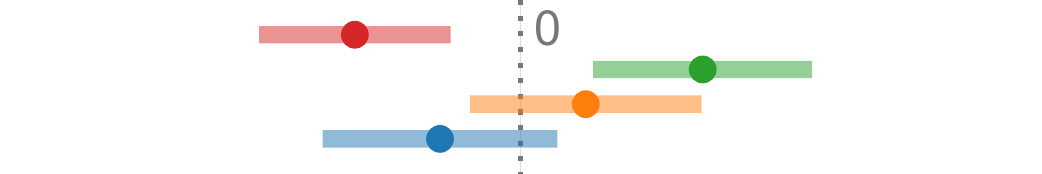}}}
\newcommand{\ProgFigCp}{\plotframe{\includegraphics[width=0.23\linewidth]{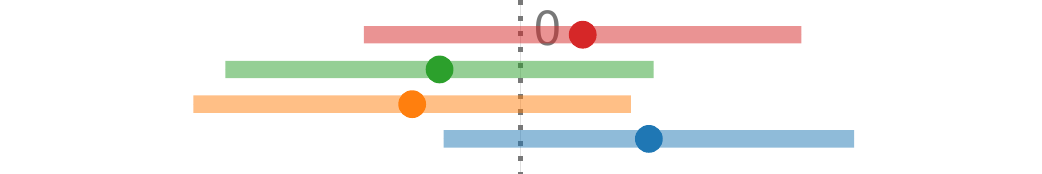}}}
\newcommand{\ProgFigDp}{\plotframe{\includegraphics[width=0.23\linewidth]{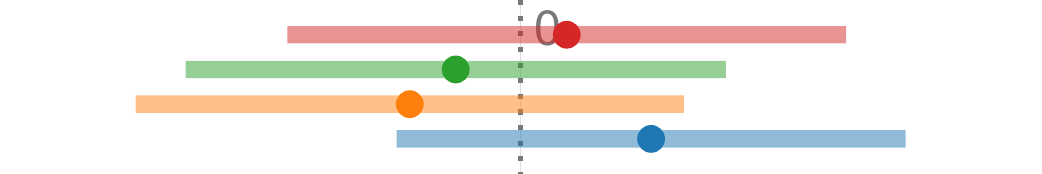}}}

\centering
\subfloat[Regular PCA]{
\begin{tikzpicture}[node distance=0.1cm]
  \node[inner sep=0pt] (a)  at (0,0)   {\ProgFigA};
  \node[inner sep=0pt, below=of a] (ap) {\ProgFigAp};
\end{tikzpicture}
}
\subfloat[\textit{s} = 0.5]{
\begin{tikzpicture}[node distance=0.1cm]
  \node[inner sep=0pt] (b) {\ProgFigB};
  \node[inner sep=0pt, below=of b] (bp) {\ProgFigBp};
\end{tikzpicture}
}
\subfloat[\textit{s} = 0.8]{
\begin{tikzpicture}[node distance=0.1cm]
  \node[inner sep=0pt] (c) {\ProgFigC};
  \node[inner sep=0pt, below=of c] (cp) {\ProgFigCp};
\end{tikzpicture}
}
\subfloat[\textit{s} = 1.0]{
\begin{tikzpicture}[node distance=0.1cm]
  \node[inner sep=0pt] (d) {\ProgFigD};
  \node[inner sep=0pt, below=of d] (dp) {\ProgFigDp};
\end{tikzpicture}
}

  \caption{%
    Data uncertainty can have a significant influence on the outcome of dimensionality reduction techniques.
    We propose a generalization of principal component analysis (PCA) that takes into account the uncertainty in the input.
    The top row shows the dataset with varying degrees of uncertainty and the corresponding principal components, whereas the bottom row shows the projection of the dataset, using our method, onto the first principal component.
    In Figures (a) and (b), with relatively low uncertainty, the blue and the orange distributions are comprised by the red and the green distributions.
    In Figures (c) and (d), with a larger amount of uncertainty, the projection changes drastically:
    now the orange and blue distributions encompass the red and the green distributions.
  }
  \label{fig:teaser}
}

\begin{document}

\firstsection{Introduction}

\maketitle

Dimensionality reduction techniques can be applied to visualize data with more than two or three dimensions, projecting the data to a lower-dimensional subspace.
These projections should be meaningful so that the important properties of the data in the high-dimensional space can still be reconstructed in the low-dimensional representation.
In general, dimensionality reduction techniques can either be linear or non-linear~\cite{Nonato2018}.
Linear dimensionality reduction has the advantage that properties and invariants of the input data are still reflected in the resulting projections.
Another advantage of linear over non-linear methods is that they are easier to reason about because the subspace for the projection is always a linear combination of the original axes.
Also, linear methods are usually efficient to implement~\cite{Cunningham2015}.

Arguably the most frequently used linear method is principal component analysis (PCA).
It is most effective if the dimensions of the input data are correlated, which is common.
PCA uses this property and finds the directions of the data that contain the largest variance.
This is achieved by performing eigenvalue decomposition of the sample covariance matrix that is estimated from the input.
The input to conventional PCA is a set of points.
However, we often encounter data afflicted with uncertainty or variances.
According to Skeels et.~\cite{Skeels2009}, there are several sources for this uncertainty.
Measurement errors might arise from imperfect observations.
In many cases, we rely on data that is the output of predictive models or simulations that provide probabilistic estimations.
And lastly, uncertainty is inevitable when aggregating data, as some of the original information has to be discarded.
The natural way to model these instances of uncertain data is by using probability distributions over possible realizations of the data.

In this paper, we derive a generalization of PCA that directly works on probability distributions.
Like regular PCA, our new method of \emph{uncertainty-aware PCA} solely requires that the expected value and the covariance between dimensions of these distributions can be determined---no higher-order statistics are taken into account.
This uncertainty of the input data can have a strong impact on the resulting projection, because it directly influences the magnitude of the eigenvalues of the sample covariance matrix.

In addition to extending PCA, we introduce \emph{factor traces} as a visualization that shows how the projections of the original axes onto the subspace change with a varying degree of uncertainty.
This enables to perform a sensitivity analysis of the dimensionality reduction with respect to uncertainty and gives an interpretable representation of the linear projection that is performed.
Our paper has four main contributions:
\begin{itemize}
  \setlength\itemsep{0em}
  \item a closed-form generalization of PCA for uncertain data,
  \item sensitivity analysis of PCA with regards to uncertainty in the data,
  \item factor traces as a new visualization technique for the sensitivity of linear projections, and
  \item establishing a distance metric between principal components.
\end{itemize}

In Figure~\ref{fig:teaser}, we compare our method to regular PCA and illustrate why it is important to consider the uncertainty in the data when determining the projection:
it shows a projection of four bivariate probability distributions, each with varying levels of uncertainty, that are projected onto a single dimension.
For input with low uncertainty, the red and green data points define the extent of the projected data.
With increasing uncertainty and due to the shape of the underlying distributions, the projection looks quite different: now the orange and blue data points mark the extent of the projected data.
This change in the projection shows that it is important to incorporate the uncertainty information adequately into our dimensionality reduction algorithms.
Although all distributions in this example are Gaussian, our method works on any probability distribution for which the expected value and the covariance can be determined.


\section{Related Work}\label{sec:related-work}

The survey by Nonato and Aupetit~\cite{Nonato2018} offers a broad overview of dimensionality reduction from a visualization perspective.
Principal component analysis~\cite{Pearson1901} is one of the oldest and most popular techniques.
It is often applied to reduce data complexity, which is a common task in visualization.
By construction, PCA yields the linear projection that retains the most variance of the input data in the lower-dimensional subspace.
Probabilistic PCA~\cite{Tipping1999} extends traditional PCA by adding a probabilistic distribution model.
In contrast to our method, an unknown isometric measurement error is assumed.
Likewise, many extensions have been introduced to PCA~\cite{Koren2004,Burges2009}.
For example, Kernel PCA~\cite{Schoelkopf1997} enables non-linear projections by first transforming objects into a higher-dimensional space in which a good linear projection can be found.
Techniques such as Bayesian PCA~\cite{Bishop1999,Nounou2002,Nakajima2011}, and the method introduced by Sanguinetti et al.~\cite{Sanguinetti2005} focus on estimating the dimensionality of the lower-dimensional space.
Robust PCA methods~\cite{Ahn2003,Tripathi2008,Vaswani2018} target datasets with outliers.
Different extensions to PCA have also been developed in the context of fuzzy systems.
The technique described by Denoeux and Masson~\cite{Denoeux2004} applies PCA to fuzzy numbers by training an artificial neural network that incorporates the different possible realizations for each fuzzy number.
Giordani and Kiers~\cite{Giordani2006} provide an overview of methods that can be used to apply PCA to interval data.
In contrast, we extend traditional PCA to an uncertainty-aware linear technique for exploratory visualization that works on general probability distributions.

Next to PCA, Factor Analysis (FA)~\cite{Spearman1904} is a well known linear method.
Its goal is to identify (not necessarily orthogonal) latent variables underlying a higher-dimensional space of measurements.
Factor Analysis models measurement errors, yet constraining the errors to be uncorrelated is common.
One reason for this is that modeling correlated errors can be problematic if the actual errors are unknown~\cite{Hermida2015}.
In our description, we assume that all errors are known, or can at least be estimated.
Many other linear techniques such as Classical Multi-Dimensional Scaling~\cite{Torgerson1952} and Independent Component Analysis~\cite{Hyvarinen2001} are covered by Cunningham and Ghahramani~\cite{Cunningham2015}.
To the best of our knowledge, none of them can deal with data that has explicitly encoded (measurement) errors.

Liu et al.~\cite{Liu2017} provide an overview of the visualization and exploration of high-dimensional data.
The Star Coordinates~\cite{Kandogan2000} visualization technique, for example, provides interactive linear projections of high-dimensional data.
Recently measure-driven approaches for exploration have gained interest, e.g., by Liu et al.~\cite{Liu2016} as well as by Lehmann and Theisel~\cite{Lehmann2016}.
Visualizing the projection matrix of linear dimensionality reduction techniques (instead of projections of the data) can be done with factor maps or Hinton diagrams~\cite{Hinton1991,Bishop1999}.

Advances in visualizing uncertainty and errors often originate from the need to represent prediction results~\cite{Spiegelhalter2011}.
More generally, visualizing Gaussian distributions by a set of isolines is a common practice.
In this paper, we aim at bringing uncertainty-aware dimensionality reduction and visualization together.
For example, our technique can be used to extend Wang et al.'s~\cite{Wang2017} approach to visualizing large datasets by allowing a fast approximate visualization of clusters.
Furthermore, correlated probability distributions are often the result of Bayesian inference, which is widely used in prediction tasks, where the result is always a probability distribution.
In this domain, Gaussian processes~\cite{Rasmussen2005} are a prime example of correlated uncertainty.

Lately, there has been a push in the visualization community to gain a better understanding of the intrinsic properties of projection methods.
However, the focus mainly has been on exploring non-linear approaches. 
For instance, Schulz et al.~\cite{Schulz2017} propose a projection for uncertainty networks based on sampling different realizations of the data and investigate potential effects of uncertainty.
With \emph{DimReader}, Faust et al.~\cite{Faust2019} address the problem of explaining non-linear projections.
Their technique uses automatic differentiation to derive a scalar field that encodes the sensitivity of the projected points against perturbations.
Wattenberg et al.~\cite{Wattenberg2016} examine how the choice of parameters affects the projection results of t-SNE.
Similarly, Streeb et al.~\cite{Streeb2018} compare a sample of (non-)linear techniques and influences of their parameters on projections.


\section{Statistical Background}\label{sec:stat-background}

The typical way to model uncertainty is by using probability distributions over the data domain.
This approach is well established in other fields, such as measurement theory and Bayesian statistics.
Before getting to the gist of our method, we want to give a quick overview of the statistical background we need for our technique.
More details can be found in the textbook by Wickens~\cite{Wickens1994}.

\subsection{Random Variables and Random Vectors}\label{sec:rand-vec}

A \emph{random variable} is used to describe the values of possible outcomes $x$ of a random phenomenon.
It is usually defined as a real-valued scalar $x \in \Real$.
Probability distributions are used to assign a probability (density) to each outcome of the random variable---both concepts are closely tied together.
To extend this one-dimensional case to multi-dimensional phenomena, we can group several random variables into a multivariate random variable, which is also called a \emph{random vector}.
Throughout this article, we denote random vectors by $\Rvec{x} = \Transpose{(x_1, \dots, x_d)}$, with $\Rvec{x} \in \Real^d$.
Analogously, the corresponding multivariate probability distributions span the same $d$-dimensional domain.
An interesting property arises from the fact that $\Rvec{x}$ can be viewed as a position vector:
it can be manipulated using affine transformations.
These transformations can, for example, be used to scale, translate, or rotate $\Rvec{x}$.
Generally, an affine transformation has the form $\Rvec{y} = \Mat{A}\Rvec{x} + \Vec{b}$.
It consists of a linear transformation $\Mat{A}$ and a translation vector $\Vec{b}$ that together transform an input $\Rvec{x}$ to obtain a new random vector $\Rvec{y}$, which can be described using a modified distribution.

\subsection{Summary Statistics}

For many applications, it is helpful to summarize the probability distributions into simpler, yet characteristic quantities.
Ideally, these simple terms still allow us to make statements about the shape and properties of the original distribution.
Such descriptions are called summary statistics.
The most well-known statistics are the first and second moments, which, in the real-valued case, are also called \emph{mean} and \emph{variance}.
They are used to describe the center of gravity and the spread of a distribution.
For multi-dimensional data, the mean is a $d$-dimensional vector, and the variance is replaced by the \emph{covariance} that also reflects correlations between each of the $d$ components.
Because the covariance describes these relationships, it has the form of a symmetric $d \times d$ matrix.
Every covariance matrix is always positive semi-definite---we provide a detailed discussion in the appendix.

For some distributions, these two summary statistics are explicitly defined.
The multivariate normal (MVN) distribution, which is widely used in many domains, has an interesting property---it is completely determined by its first and second moments.
Therefore, if $\Rvec{x}$ follows an MVN distribution, with mean $\DistMean$ and covariance matrix $\DistCov$, we write:
$$
\Rvec{x} \Follows \Normal(\DistMean, \DistCov).
$$

Sometimes our random vector $\Rvec{x}$ is given by a set of samples $\{\Vec{x}_n\},\, n \in \{1,\dots,N\}$ from an arbitrary distribution.
Given this set, we can estimate the first and second moments of $\Rvec{x}$ using the sample mean $\Mean{\Rvec{x}}$, which is defined in terms of the expected value $\Ev{\cdot}$:
$$
\Mean{\Rvec{x}} = \Ev{\Rvec{x}} = \frac{1}{N}\sum^N_{n=1}\Vec{x}_n
$$
and the sample covariance matrix $\Cov{\Rvec{x}}{\Rvec{x}}$:
\begin{align}
  \Cov{\Rvec{x}}{\Rvec{x}}
  &= \Ev{\OuterS{(\Rvec{x} - \Ev{\Rvec{x}})}} \nonumber\\
  &= \Ev{\OuterS{\Rvec{x}}} - \OuterS{\Mean{\Rvec{x}}} \label{eq:cov}
\end{align}
The term $\Ev{\OuterS{\Rvec{x}}}$ is the expected outer product $\OuterS{\Rvec{x}}$ and can be approximated as follows:
$$
\Ev{\OuterS{\Rvec{x}}} = \frac{1}{N}\sum^N_{n=1}\OuterS{\Vec{x}_n}
$$

In the previous section, we explained how to transform a random vector $\Rvec{x}$ using affine transformations.
Transforming $\Rvec{x}$ in this way also influences the summary statistics.
For the mean, it holds that:
\begin{align}
 \Ev{\Mat{A}\Rvec{x} + \Vec{b}} = \Mat{A}\Ev{\Rvec{x}} + \Vec{b} \nonumber
\end{align}
In a similar fashion, we can transform the covariance matrix:
\begin{align}
  \Cov{\Mat{A}\Rvec{x} + \Vec{b}}{\Mat{A}\Rvec{x} + \Vec{b}} = \Mat{A}\Cov{\Rvec{x}}{\Rvec{x}}\Transpose{\Mat{A}} \label{eq:affine-cov}
\end{align}
Both equations follow from the linearity of the expected value operator $\Ev{\cdot}$.
Intuitively, only the mean of $\Rvec{x}$ is influenced by the translation $\Vec{b}$.
The covariance matrix, in contrast, is invariant to translation.
The reason for this is that the covariance only captures the relative variance of each component because it is always centered around the sampling mean by the term $\OuterS{\Mean{\Rvec{x}}}$.
In the following section, we will use these above definitions to formulate our method.


\section{Method}\label{sec:method}

We have motivated the different causes of uncertainty in the input data in the introduction.
In this part, we describe the necessary adaptions to the framework of PCA that are required to handle uncertainty, as modeled in the previous section.
We will first show how to adapt the computation of the covariance matrix to work on probability distributions, which is a fundamental part of our technique.
Then, we will describe how this fits into the context of regular PCA.
Afterward, we will demonstrate how our method allows us to perform PCA analytically on uncertain data, using multivariate normal distributions as an example.
Finally, we will show that our approach is a generalization of regular PCA.
This allows us to combine certain and uncertain data within the same mathematical framework and provides us with the foundation for sensitivity analysis, as described in Section~\ref{sec:sensitivity-analysis}.

\subsection{Model}
PCA is used to find the directions of the data with the largest variance by looking at the covariance of the input.
We adopt this concept to arbitrary distributions to handle uncertain data.
For our method, we only require that the expected value and the covariance can be determined for each of the distributions.
It is important to note that this does not imply that the input distributions necessarily have to follow a Gaussian distribution.
We want to illustrate this for a small example:
let us consider an input distribution made up of two clusters spread about its mean.
Then, the covariance of the distribution still captures the spread of the data, namely along the direction of the location of the two clusters.
So even though the distribution might not be sufficiently described only by mean and covariance, its overall extent is still represented adequately using these first- and second-order statistics.
In Section~\ref{sec:anuran}, we will show an example of a dataset that exhibits this property.
And in Section~\ref{sec:limitations}, we will discuss its implications on the resulting projection.

It is important that there is an established relationship between the units of the original axes for PCA to yield a meaningful result.
The usual approach to achieve this is to normalize the input data accordingly.
The same preprocessing step needs to be performed for our method.
For probability distributions, this can be performed using affine transformations, as outlined above.

\subsection{Uncertain Covariance Matrices}\label{sec:uncertain-covariances}

As we have mentioned before, the goal of our method is to perform PCA on a set of $N$ probability distributions that are used to model the uncertainty, as described in Section~\ref{sec:stat-background}.
Formally, we represent this collection of distributions as random vectors $\ObSet = \{\Ob_1, \dots, \Ob_N\}$.
For each of these random vectors, we require that we can determine its expected value $\Ev{\Ob_i}$ and its pairwise covariance $\Cov{\Ob_i}{\Ob_i}$.
It is important to note that $\ObSet$ can conceptually be interpreted as a random vector of second order, as its components $\Ob_n$ are random vectors themselves.

Our approach adapts the computation of the covariance matrix to account for uncertainty in the data.
Regular PCA works on a set of points. Therefore, the covariance matrix can be understood as the computation of the expected products of deviations of these points from the sample mean.
In contrast, our approach works on a set of random vectors, which changes the problem in the following way:
Because of the uncertainty in the data, we do not know the actual deviation of each random vector from the overall sample mean.
But we can determine the deviation that is to be expected for each of the distributions.
We do this conceptually by integrating over the deviation of all possible realizations of each probability distribution.
In the framework of PCA, where only the first- and second-order moments are taken into account, it turns out we do not even have to evaluate this integral:
we can derive the covariance matrix directly from the summary statistics.

From Equation~\ref{eq:cov}, we can derive a property of the covariance matrix that we will need later on:
it gives us a way to compute the expected outer product $\Ev{\OuterS{\Rvec{x}}}$ of a particular random vector with itself.
We achieve this by solving Equation~\ref{eq:cov} for $\Ev{\OuterS{\Rvec{x}}}$:
\begin{align}
  \Ev{\OuterS{\Rvec{x}}} = \OuterS{\Ev{\Rvec{x}}} + \Cov{\Rvec{x}}{\Rvec{x}} \label{eq:solve}
\end{align}

For distributions, we use the following equation, which is akin to computing the expected products of \emph{expected} deviations.
To avoid confusion with the expected value of each random vector $\EvOb{\cdot}$, we denote the expectation operator that stems from the covariance method with $\EvCov{\cdot}$:
$$
\Cov{\ObSet}{\ObSet} = \EvCov{\EvOb{\OuterS{\ObSet}} - \Centering{\ObSet}}
$$
We can expand this further by making use of Equation~\ref{eq:solve}:
\begin{align}
  \Cov{\ObSet}{\ObSet}
  &= \EvCov{\OuterS{\EvOb{\ObSet}} + \Cov{\ObSet}{\ObSet} - \Centering{\ObSet}} \nonumber
\end{align}
\eqframe{eq:final}{\Cov{\ObSet}{\ObSet}=\EvCov{\OuterS{\EvOb{\ObSet}}} + \EvCov{\Cov{\ObSet}{\ObSet}} - \Centering{\ObSet}}

The different terms in Equation~\ref{eq:final} have particular interpretations.
First, we recognize that the term $\EvCov{\OuterS{\EvOb{\ObSet}}}$ is the same as performing regular PCA on the means of each of the distributions.
The second term $\EvCov{\Cov{\ObSet}{\ObSet}}$ computes the average covariance matrix over all random vectors:
\begin{align}
  \EvCov{\Cov{\ObSet}{\ObSet}} = \frac{1}{N} \sum_{i=1}^{N} \Cov{\Ob_i}{\Ob_i} \label{eq:expected-cov}
\end{align}
It reflects the uncertainty that each random vector has and how these uncertainties influence the overall covariance in the dataset---it is also the major difference between our method and regular PCA, which cannot handle probability distributions.
The last term is called centering matrix and also part of regular PCA.
It consists of the outer product of the empirical mean $\Mean{\ObSet}$ of our dataset.
The empirical mean of our dataset can be computed as follows:
$$
\Mean{\ObSet} = \frac{1}{N} \sum_{i=1}^{N} \EvOb{\Ob_i}
$$

Algorithm~\ref{alg:covariance} provides the corresponding pseudocode for Equation~\ref{eq:final}.
The proof that Equation~\ref{eq:final} yields a symmetric, positive semi-definite matrix and therefore is an actual covariance matrix can be found in the Appendix of this document.

\begin{algorithm}[!b]
  \label{alg:covariance}
  \DontPrintSemicolon
  \SetKwInOut{Input}{Input}
  \SetKwInOut{Output}{Output}
  \caption{Covariance matrix of random vectors}
  \Input{ List of $d$-variate distributions $\ObSet$, scaling factor $s=1$}
  \Output{ Covariance matrix $\Mat{K}_{\ObSet\ObSet}$}
  \BlankLine
  $\Mean{\Ob} \leftarrow d$-dimensional vector initialized to $0$\;
  \ForEach{$\Ob \in \ObSet$}{
    $\Mean{\Ob} \mathrel{+}= \Ob.mean()$\;
  }
  $\Mean{\Ob} \mathrel{\slash}= \ObSet.length()$\;
  $\Mat{K}_{\ObSet\ObSet} \leftarrow d \times d$ matrix initialized to $0$\;
  \ForEach{$\Ob \in \ObSet$}{
    $\Vec{m} \leftarrow \Ob.mean()$\;
    $\Mat{K}_{\ObSet\ObSet} \mathrel{+}= \OuterS{\Vec{m}} + s^2 \cdot \Rvec{t}.cov() - \Centering{\Ob}$\;
  }
  $\Mat{K}_{\ObSet\ObSet} \mathrel{\slash}= \ObSet.length()$\;
  \Return $K_{\ObSet\ObSet}$\;
\end{algorithm}

\subsection{PCA Framework and Diagonalization}

Now that we have constructed the covariance matrix while respecting the uncertainty, we can continue with the remaining steps of the PCA algorithm.
After setting up the covariance matrix, we retrieve its eigenvalues $\Eigenvalue_d$ and corresponding eigenvectors $\Eigenvector_d$.
This can be done using \emph{eigenvalue decomposition}:
$$
\Cov{\ObSet}{\ObSet}\Eigenvector = \Eigenvalue\Eigenvector
$$
Let $q$ be the desired target number of dimension for our dimensionality reduction.
We then choose the $q$ largest $\Eigenvector_d$ by their corresponding eigenvalue $\Eigenvalue_d$, yielding $q$ principal components $\PrincipalCompSet = \{\PrincipalComp_1,\dots,\PrincipalComp_q\}$.
We can then project each distribution onto the subspace $\Span{\PrincipalCompSet}$ that is spanned by these principal components $\Proj{\Ob_n} \in \Span{\PrincipalCompSet}$, where $\Proj{\cdot}$ is a linear projection that can be described using a linear transformation.

It is important to note that eigenvalues and eigenvectors have certain characteristics that complicate their analysis.
The orientation of $\Eigenvector_d$ is not completely defined, therefore $\Eigenvector_d \hat{=} -\Eigenvector_d$.
In practice, the computation of $(\Eigenvalue_d, \Eigenvector_d)$ is performed numerically, which can lead to instabilities and rounding errors.
We will discuss the impact of this on the analysis of linear projections in Section~\ref{sec:sensitivity-analysis}.

\subsection{Linear Transformation of MVN Distributions}

Now that we have defined the projection $\Proj{\cdot}$, we need to transform each distribution into the subspace $\Span{\PrincipalCompSet}$.
In the following, we will describe how this can be carried out for multivariate normal distributions, as they are often used to model errors or uncertainty in the data.
As mentioned in Section~\ref{sec:related-work}, several existing techniques already model uncertainty using MVN distributions.
In these works, the distributions are usually described using an error model, which means that a measurement $\Vec{x}$ is disturbed by an error term $\Error$.
This is commonly written as:
$$
\Ob_n = \Vec{x}_n + \Rvec{\Error}_n, \quad \Rvec{\Error}_n \Follows \Normal(\Zero, \DistCov_n)
$$
To retrace the closed-form derivation of the covariance matrix that we described in Section~\ref{sec:uncertain-covariances}, it is easier to think of this error in terms of a single random vector $\Ob_n$ that can be equivalently defined as follows:
$$
\Ob_n \Follows \Normal(\Vec{x}_n, \DistCov_n).
$$
\begin{figure}[tb]
  \centering
  \subfloat[]{\plotframe{\includegraphics[width=0.313\linewidth]{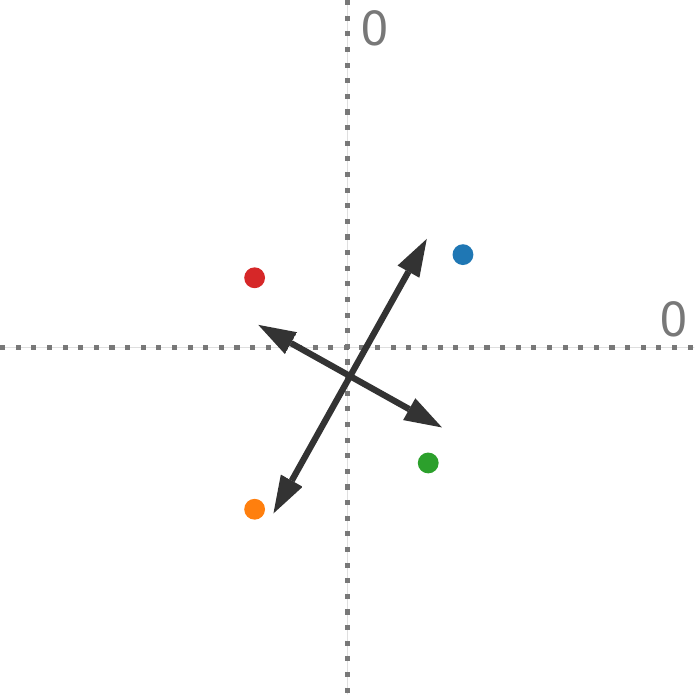}}}\,
  \subfloat[]{\plotframe{\includegraphics[width=0.313\linewidth]{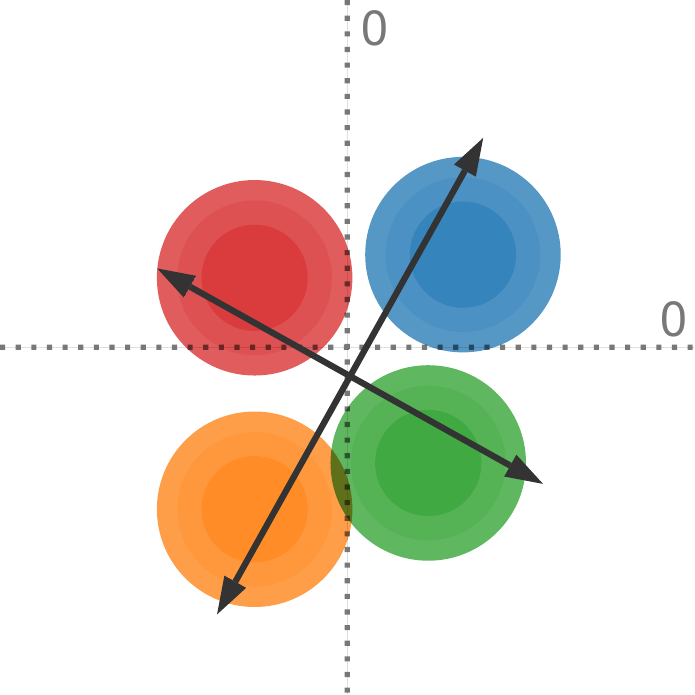}}}\,
  \subfloat[]{\plotframe{\includegraphics[width=0.313\linewidth]{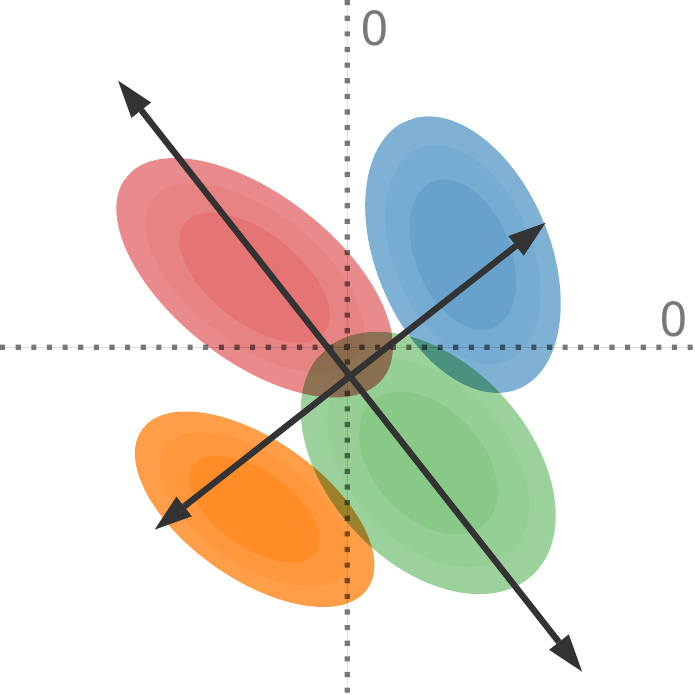}}}
  \caption{%
    Different types of input data:
    (a) Regular PCA without uncertainty.
    (b) Isometric error model as used by previous work where PCA has been described as an optimization problem;
    the directions of the principal components are the same, but the lengths differ.
    (c) Our method: it works on arbitrary distributions and can result in drastically different principal components.
  }
  \label{fig:distribution-types}
\end{figure}%
Figure~\ref{fig:distribution-types} shows examples of different error models that can be created depending on the shape of $\DistCov_n$.
We also visualize the corresponding principal components of the dataset, determined by using our method.

The dimensionality of $\Proj{\Ob}$ is $\Dim(\Span{\PrincipalCompSet})$.
To perform the actual projection, we assume that $\PrincipalComp_{q}$ are unit vectors, and write them in a column matrix $\ProjColumnMat$:
$$
\ProjColumnMat = \left[\PrincipalComp_1\,\dots\,\PrincipalComp_q\right]
$$
It is important to note that a projection $\ProjOper$ is an affine transformation, as defined in Section~\ref{sec:stat-background}.
Accordingly, we can project a normal distribution as follows:
$$
\Proj{\Ob_n} = \Normal(\Transpose{\ProjColumnMat} \DistMean_n,\: \Transpose{\ProjColumnMat} \DistCov_n \ProjColumnMat)
$$
The resulting distribution remains multivariate normally distributed.

\begin{figure}[bt]
  \centering
  \captionsetup[subfigure]{labelformat=empty,font={sf,footnotesize}}
  \subfloat[High-dimensional space]{\plotframe{\includegraphics[width=0.483\linewidth]{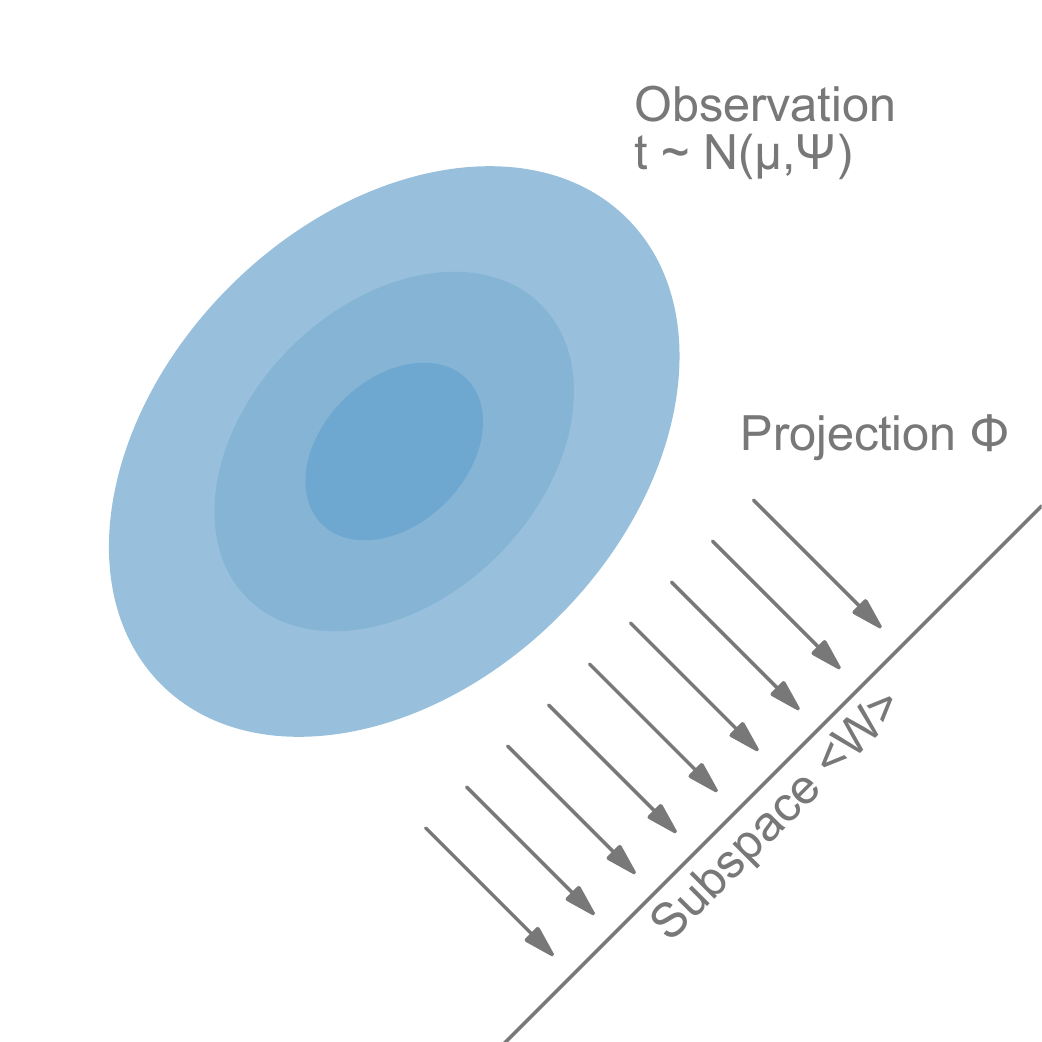}}}\,
  \subfloat[Low-dimensional subspace]{\plotframe{\includegraphics[width=0.483\linewidth]{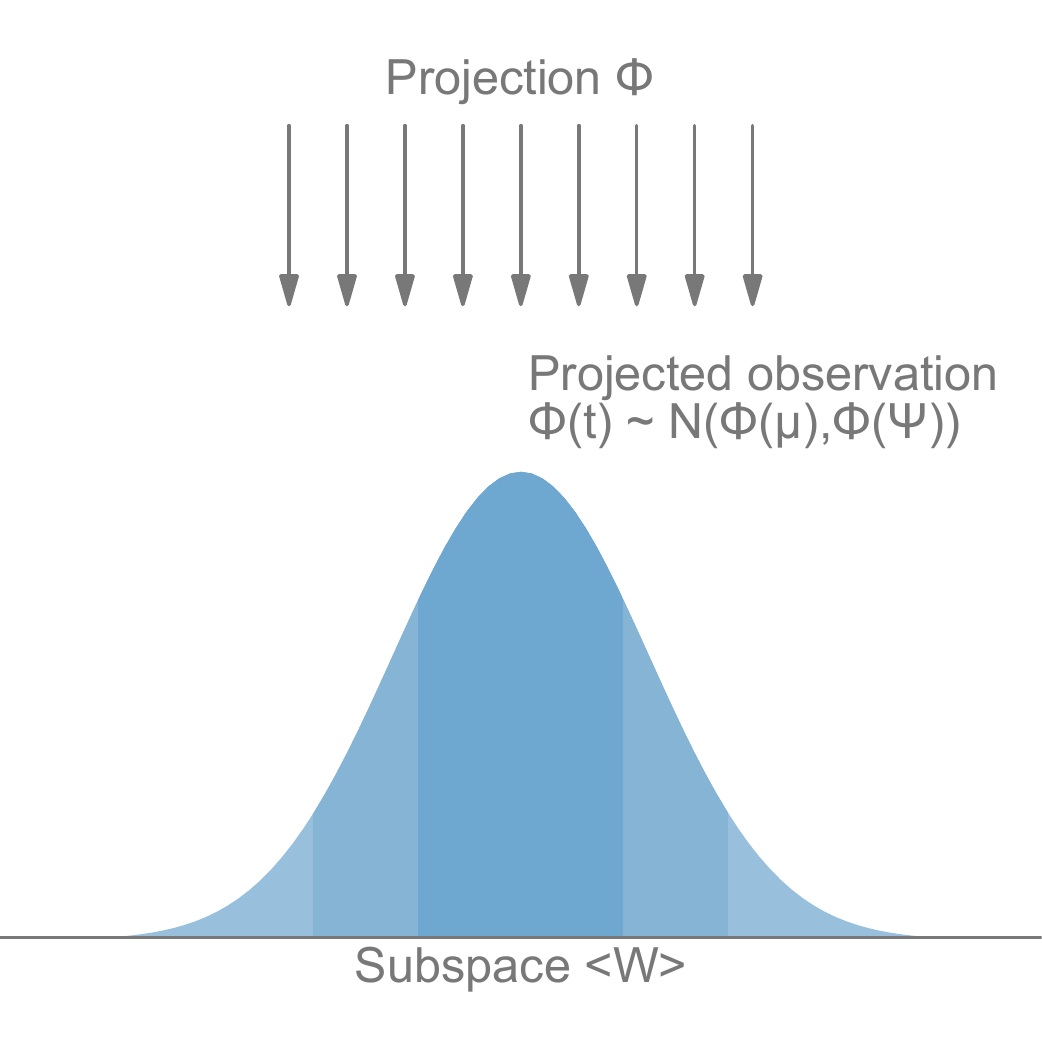}}}
  \caption{%
    A linear projection $\Proj{\cdot}$ of a normal distribution $\Ob \Follows \Normal(\DistMean, \DistCov)$ results in a modified multivariate normal distribution $\Proj{\Ob}$ in the lower dimensional subspace $\Span{\PrincipalCompSet}$.
    Because of this, we can propagate the uncertainty directly through linear dimensionality reduction techniques.
  }
  \label{fig:closed-projection}
\end{figure}

\subsection{Reduction to Regular PCA}\label{sec:generalization}

In this section, we will show that our method is a mathematical generalization of conventional PCA.
The main difference between the two algorithms lies in the setup of the covariance matrix, as described by Equation~\ref{eq:final}.
Our method includes an additional term that reflects the uncertainty of each input (Equation~\ref{eq:expected-cov}).
To reduce our formulation to regular PCA, we will scale the covariance of each of the distributions by a constant factor $s$.
This decreases the spread of the covariance matrix, and because of this, implicitly reduces the amount of uncertainty within each distribution.

To scale the covariance matrices, we will again make use of the properties of affine transformations for covariance matrices, as discussed in Section~\ref{sec:stat-background}.
Let $\Mat{S}$ be a scale matrix that has the form $\Mat{S} = \Diag(s)$.
We can now use Equation~\ref{eq:affine-cov} to scale $\ExpCov = \EvCov{\Cov{\ObSet}{\ObSet}}$:
$$
\Mat{S}\left(\ExpCov\right)\Transpose{\Mat{S}}
$$
In practice, we can make use of the fact that a scale matrix $\Mat{S}$ is always a diagonal matrix.
In our case, each diagonal entry is equal to $s$, therefore $\Mat{S} = \Diag(s)$.
This allows us to simplify the equation above even further:
\eqframe{eq:scaling}{\Mat{S}\left(\ExpCov\right)\Transpose{\Mat{S}} = s^2\cdot\ExpCov}

\begin{figure}[bt]
  \centering
  \captionsetup[subfigure]{labelformat=empty}
  \subfloat[\textit{s} = 0.25]{\plotframe{\includegraphics[width=0.313\linewidth]{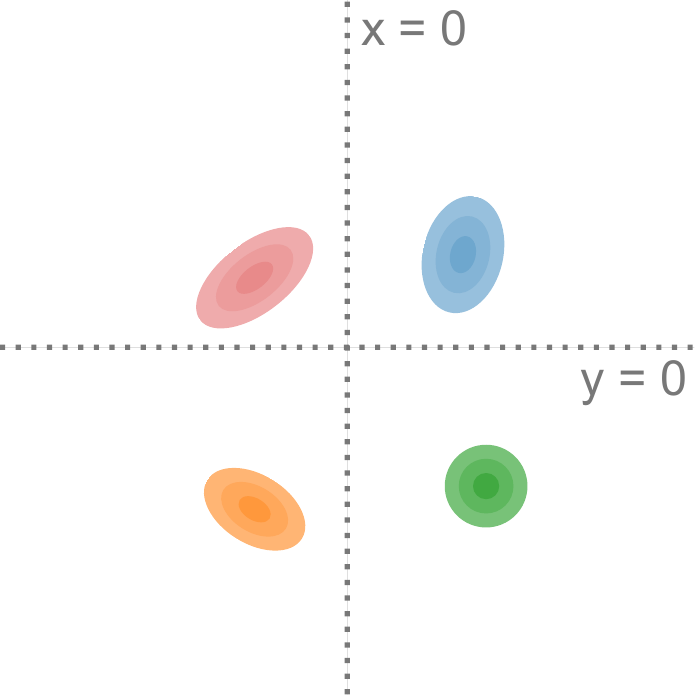}}}\,
  \subfloat[\textit{s} = 0.50]{\plotframe{\includegraphics[width=0.313\linewidth]{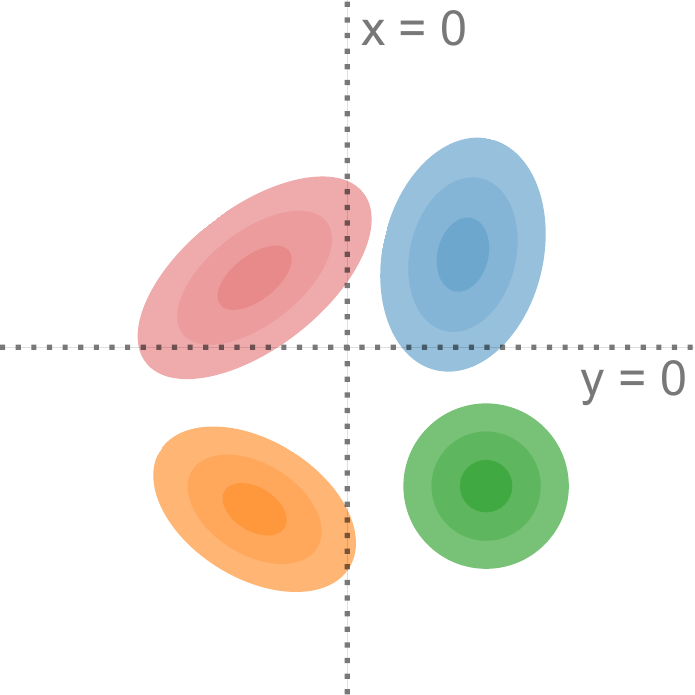}}}\,
  \subfloat[\textit{s} = 0.75]{\plotframe{\includegraphics[width=0.313\linewidth]{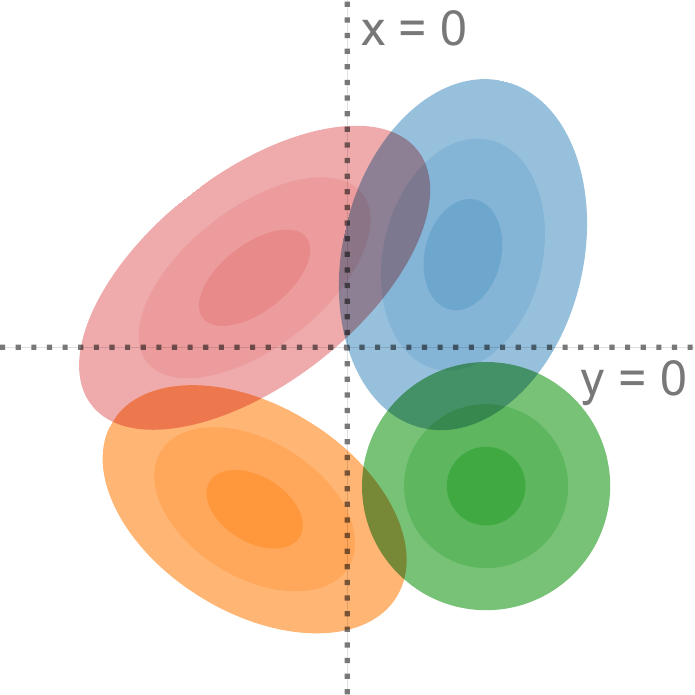}}}
  \caption{%
    Different levels of uncertainty can be achieved by scaling the covariances of each distribution with a factor \textit{s}.
    By letting \textit{s} $\mathsf{\rightarrow}$ 0 we can emulate traditional principal component analysis, as the distributions converge toward single points.
  }
  \label{fig:scaling-uncertainty}
\end{figure}

Figure~\ref{fig:scaling-uncertainty} shows a set of multivariate normal distributions, all scaled with different weights.
Another property of this description is that we can use $s$ to interpolate between the certain and uncertain representation of our data.
Algorithm~\ref{alg:covariance} shows how to incorporate the scaling factor into the computation of the covariance matrix.
In the next section, we will use this fact to investigate how much the uncertainty influences the resulting projection.

\section{Sensitivity Analysis}\label{sec:sensitivity-analysis}

We have shown in previous sections that uncertainty in the input can have a strong influence on the resulting set of principal components.
Therefore, to better understand this relationship, we investigate to what amount the dimensionality reduction depends on the shape of each of the probability distributions.
In Section~\ref{sec:generalization}, we have shown that our method is a generalized formulation of conventional PCA.
We achieved this by scaling the covariances of each distribution with a factor $s$ that describes the importance of the uncertainty.
Now, we will leverage this model to show how the fitted projection varies for different scaling factors in the interval $s \in [0, \infty)$.
This interval can be split up in two parts to investigate two different scenarios.
For $0 \leq s \leq 1$, we can interpolate between uncertainty-aware PCA and regular PCA.
Conversely, by choosing $1 < s < \infty$ we can extrapolate what the projection would look like if the uncertainty were higher.
In the following, we propose a novel visualization technique that is tailored to analyze the effects of different scaling factors $s$ and hence influences of different levels of uncertainty.

\subsection{Factor Traces}

Factor Analysis shares many similarities with PCA and is often used for the explanatory analysis of multi-dimensional datasets.
The individual latent factors, akin to principal components, are usually represented using \emph{factor maps}.
To create a factor map, the unit vectors of each dimension in feature space are projected according to the latent factors of the data~\cite{Steiger2015}.
We extend this technique to enable the exploration of the effects of uncertainty on PCA.

\begin{figure}[tb]
  \centering
  \input{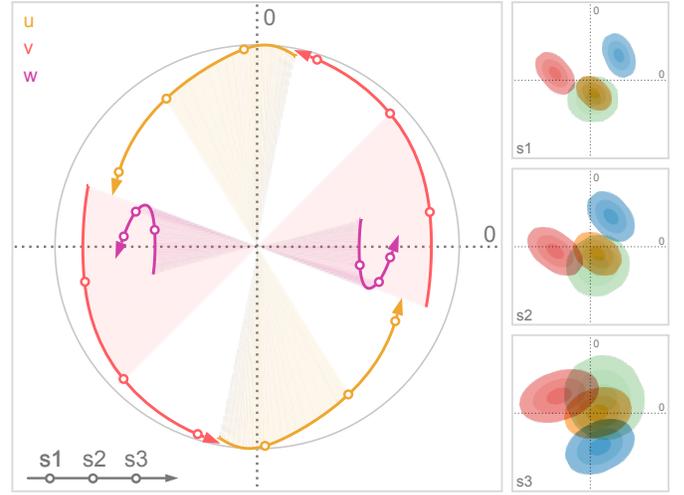}
  \caption{%
    Progression of a factor trace (left) as the uncertainty increases.
    Both unit vectors \textit{u} and \textit{v} rotate around \textit{w}.
    Accordingly, the projected distributions (right) rotate around the mean as well.
  }
  \label{fig:progression}
\end{figure}

Factor maps visualize static latent information that is hidden in the input data.
However, we are interested in visualizing the progression of uncertainty.
We do this by looking at the \emph{factor traces} that are described by the change of principal components under the varying degree of uncertainty.
In particular, we perform sensitivity analysis by continuously scaling the covariances of the distributions from the original dataset using $s$ as a scaling factor, as described above.
Figure~\ref{fig:progression} shows an example of factor traces of a three-dimensional dataset.
For each $s \in [0, \infty)$ a different subspace is chosen.
As a result, the projected unit vectors describe a trace in the image space.
Thereby, we obtain a compact representation of the analogous transformation of the feature space coordinate system.
As we mentioned before, there are two intervals for $s$ that are of interest for the analysis of the sensitivity with respect to the uncertainty.
The interval $0 \leq s \leq 1$ is highlighted by shading the area under the trace.
In contrast, for the interval $1 < s < \infty$ we only show the trace to avoid visual clutter, and we use an arrowhead to represent $s \rightarrow \infty$.

In practice, we progressively sample $s$ in the interval using a hyperbolic function.
At the heart of principal component analysis is the decomposition of the covariance matrix into its eigenvalues and eigenvectors.
This entails various challenges for the interpretation of the projection.
While the eigenvectors of a positive semi-definite matrix are always orthogonal to each other, their orientation is ambiguous as their sign can change.
In the resulting sequence, it can happen that the sign of $\Vec{v}_i$ and $\Vec{v}_{i+1}$ flips.
This, in return, leads to a mirrored projection.
We account for this in factor traces by providing both projections of the unit vectors of the original axes.
For example, this becomes apparent when looking at the purple trace in Figure~\ref{fig:progression}.
We discuss the limitations of this approach in Section~\ref{sec:discussion}.

\subsection{Interpretation}

\begin{figure}[tb]
  \centering
  \subfloat[Factor traces (Iris)]{\plotframe{\includegraphics[width=0.483\linewidth]{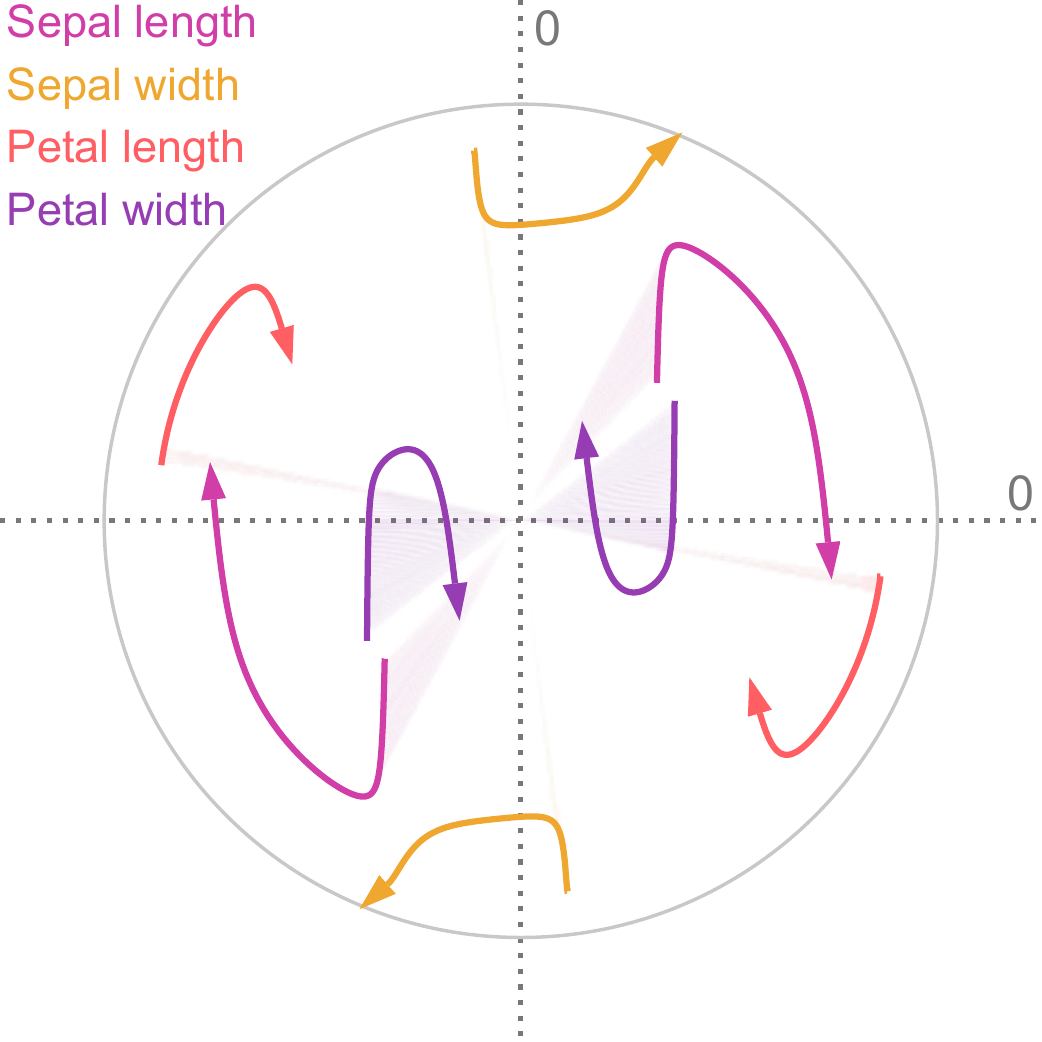}}}\,
  \subfloat[Eigenvalues (Iris)]{\plotframe{\includegraphics[width=0.483\linewidth]{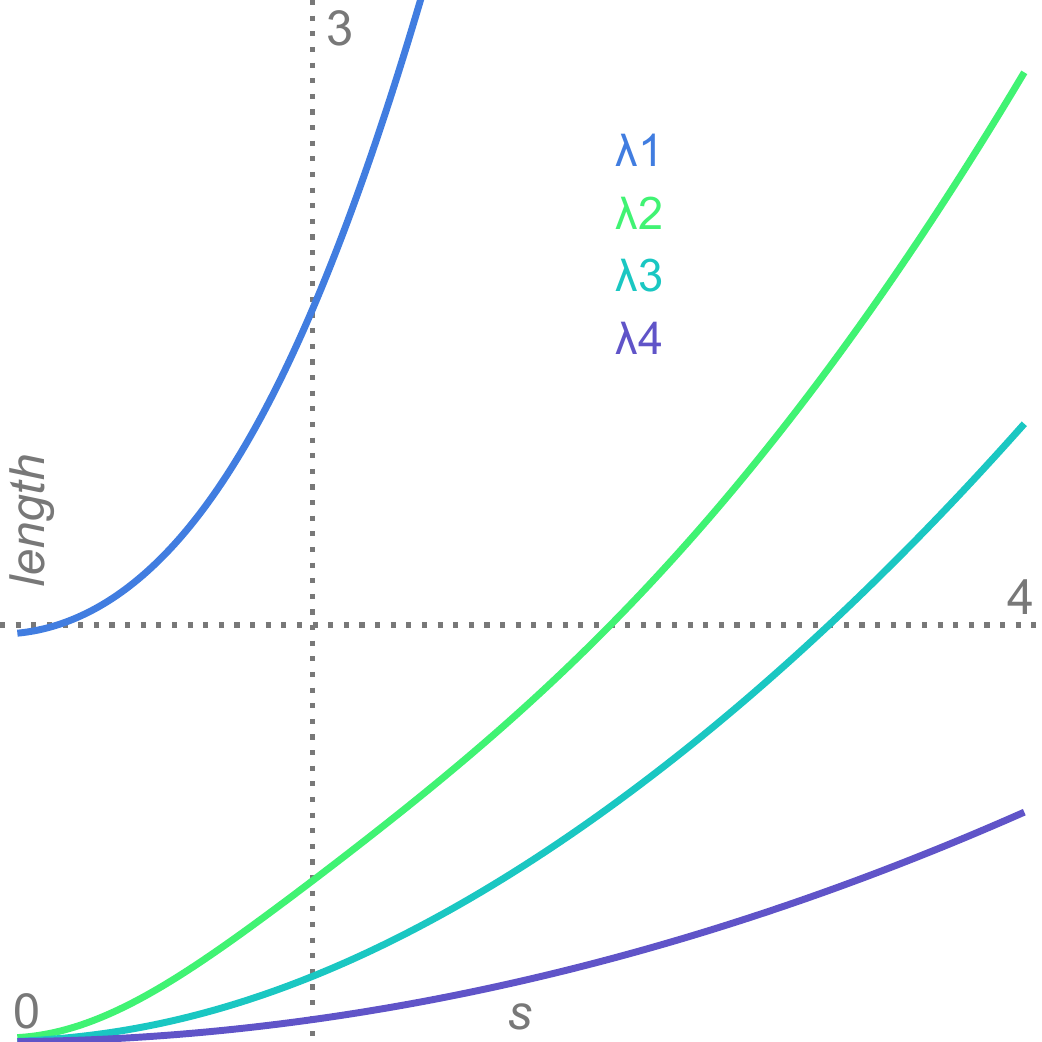}}}\\
  \subfloat[Factor traces (synth.)]{\plotframe{\includegraphics[width=0.483\linewidth]{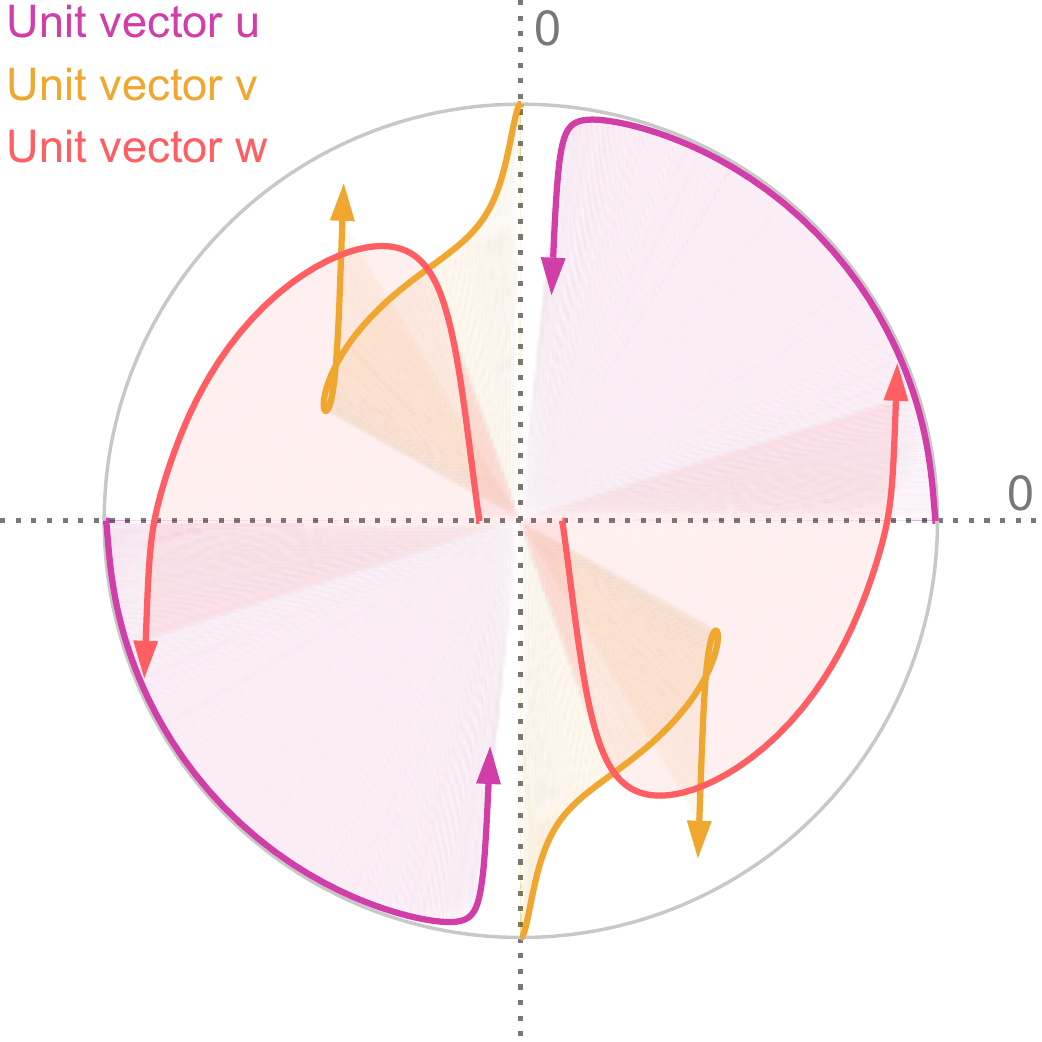}}}\,
  \subfloat[Eigenvalues (synth.)]{\plotframe{\includegraphics[width=0.483\linewidth]{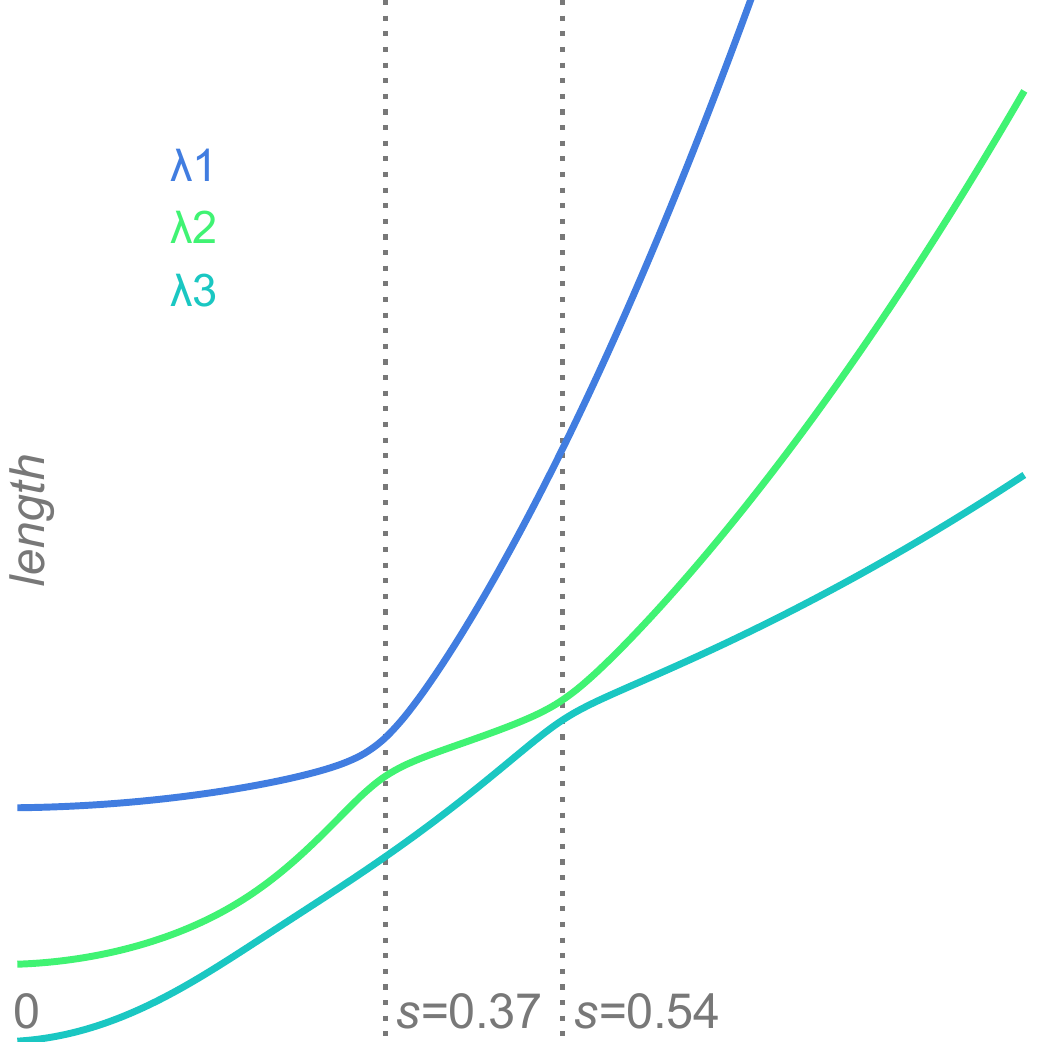}}}
  \caption{%
    Factor traces for two different datasets:
    (a) The 4D Iris dataset with (b) corresponding plot of the eigenvalues.
    (c) A 3D synthetic dataset, also with eigenvalues (d).
    Whereas the Iris dataset has no avoided crossing eigenvalues, the synthetic dataset has two avoided crossings (d) represented by bumps in the plot (\textit{s} $\in$ \{0.37,0.54\}).
    The factor traces are projected onto a 2D subspace---as a consequence only the second bump manifests in the traces:
    at \textit{s} $\approx$ 0.54 the orange trace \textit{v} forms a loop, while the purple trace \textit{u} curves inward.
  }
  \label{fig:eigenvalues}
\end{figure}

Factor traces simultaneously visualize different properties of the original dataset with respect to the corresponding projection:
the length of each trace describes how strongly each original axis is affected by the uncertainty in the data, whereas the distance of each part of the trace to the center depicts the linear combination of the original unit vectors that define the projection.
Factor traces also offer a way to analyze the robustness of the resulting projections with respect to uncertainty.
The covariance matrices and the overall shape of the data determine the corresponding eigenvalues.
Because the principal components are sorted by their eigenvalues and only the $q$ largest eigenvalues are chosen, their respective values also have a large effect on the resulting projection.
Figure~\ref{fig:eigenvalues} shows factor traces of two different datasets, together with plots of their eigenvalues.
With an increasing $s$, sometimes the distance between two eigenvalues $\lambda_i, \lambda_j$ decreases more and more.
In some cases, it appears that the eigenvalues will cross, but instead, they will eventually start to move away from each other again.
This effect closely resembles \emph{avoided crossings}, a quantum phenomenom~\cite{Neumann1991}.
The reason for this effect is that two eigenvalues coalesce as they end up with the same length~\cite{Seyranian2005}.
Eigenvalues that avoid crossing manifest in distinctive bumps in their corresponding eigenvalue plots, which can be seen in Figure~\ref{fig:eigenvalues}d.
The first dataset in Figure~\ref{fig:eigenvalues} does not contain any avoided crossings.
By contrast, Figure~\ref{fig:eigenvalues}c and Figure~\ref{fig:eigenvalues}d show a three-dimensional dataset with two bumps (highlighted by the dotted lines).
Avoided crossings make it difficult to reason about the behavior of the eigenvectors and consequently the resulting projection in these points.
In some cases---Figure~\ref{fig:eigenvalues}c, for example---we can observe sharp turns in the corresponding factor traces.
Here, the avoided crossing is between $\lambda_2$ and $\lambda_3$.

In conjunction with PCA, factor traces can aid the exploratory analysis of datasets by giving insights into the behavior of the principal components under uncertainty.
Apart from showing how the projection changes under uncertainty, factor traces can help gauge how robust and hence how trustful the projected view of the dataset is.
While our approach can aid in assessing projections, the visualization of high-dimensional data involving a large variety of distributions remains a difficult challenge.
Generally, factor traces work well for datasets with up to six original dimensions.
Above this limit, the representation becomes more difficult to understand due to overplotting. 
As shown in Figure~\ref{fig:eigenvalues}, the interpretation of factor traces can be further enhanced by taking the corresponding eigenvalue plot into account.
Depending on the dataset, we see the possibility to encode this information directly onto the factor trace, either by thickness or color.

\begin{figure}[bt]
  \centering
  \subfloat[Traditional PCA]{\plotframe{\includegraphics[width=0.483\linewidth]{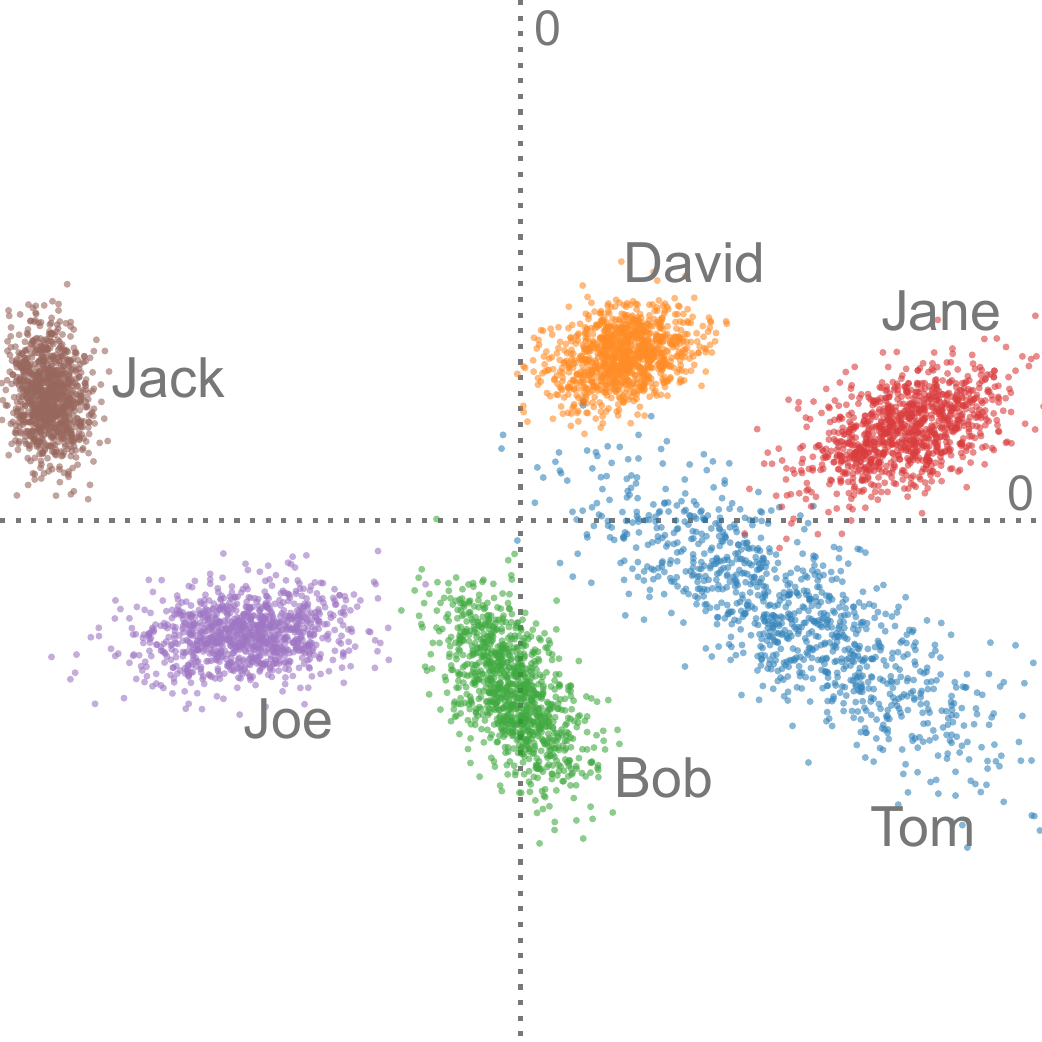}}}\,
  \subfloat[Uncertainty-aware PCA (ours)]{\plotframe{\includegraphics[width=0.483\linewidth]{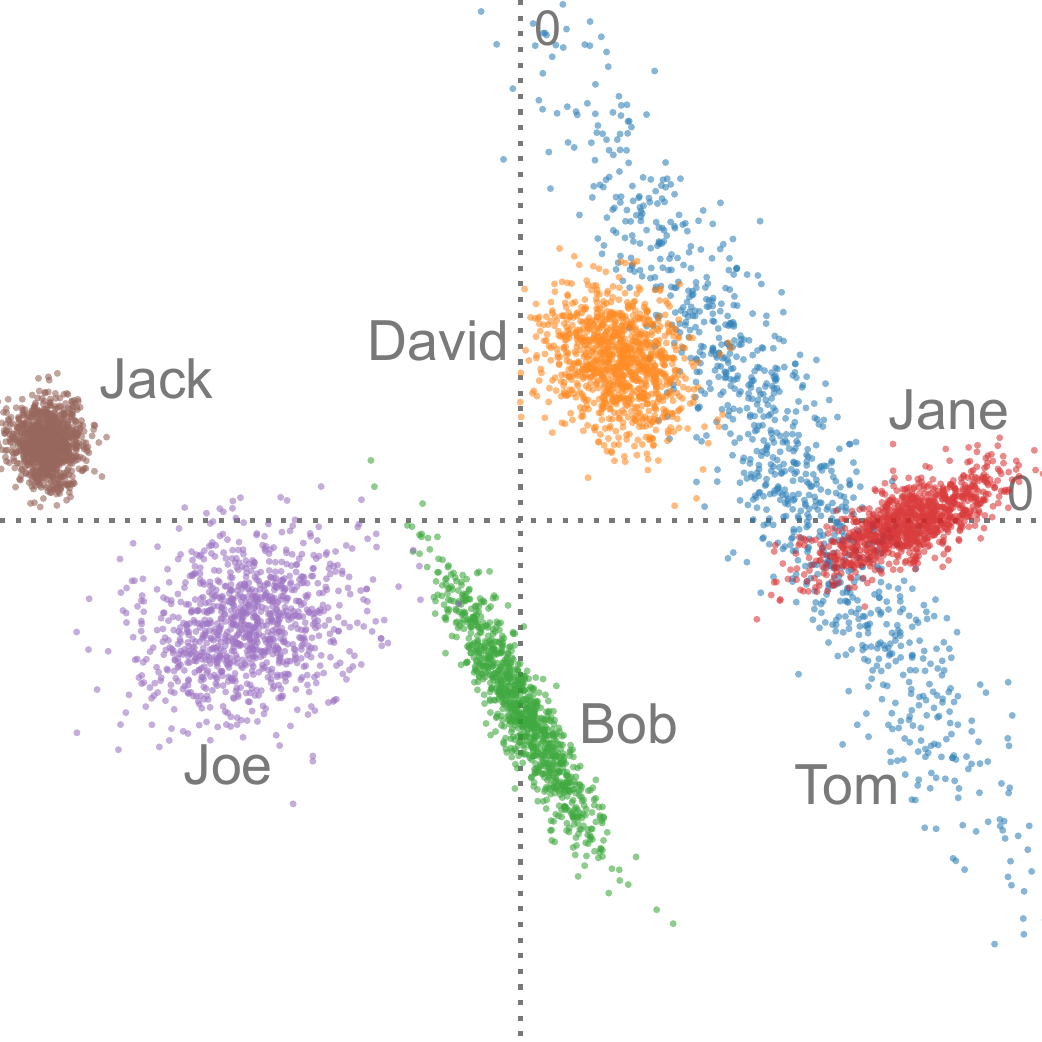}}}\\
  \subfloat[Factor traces]{\plotframe{\includegraphics[width=0.483\linewidth]{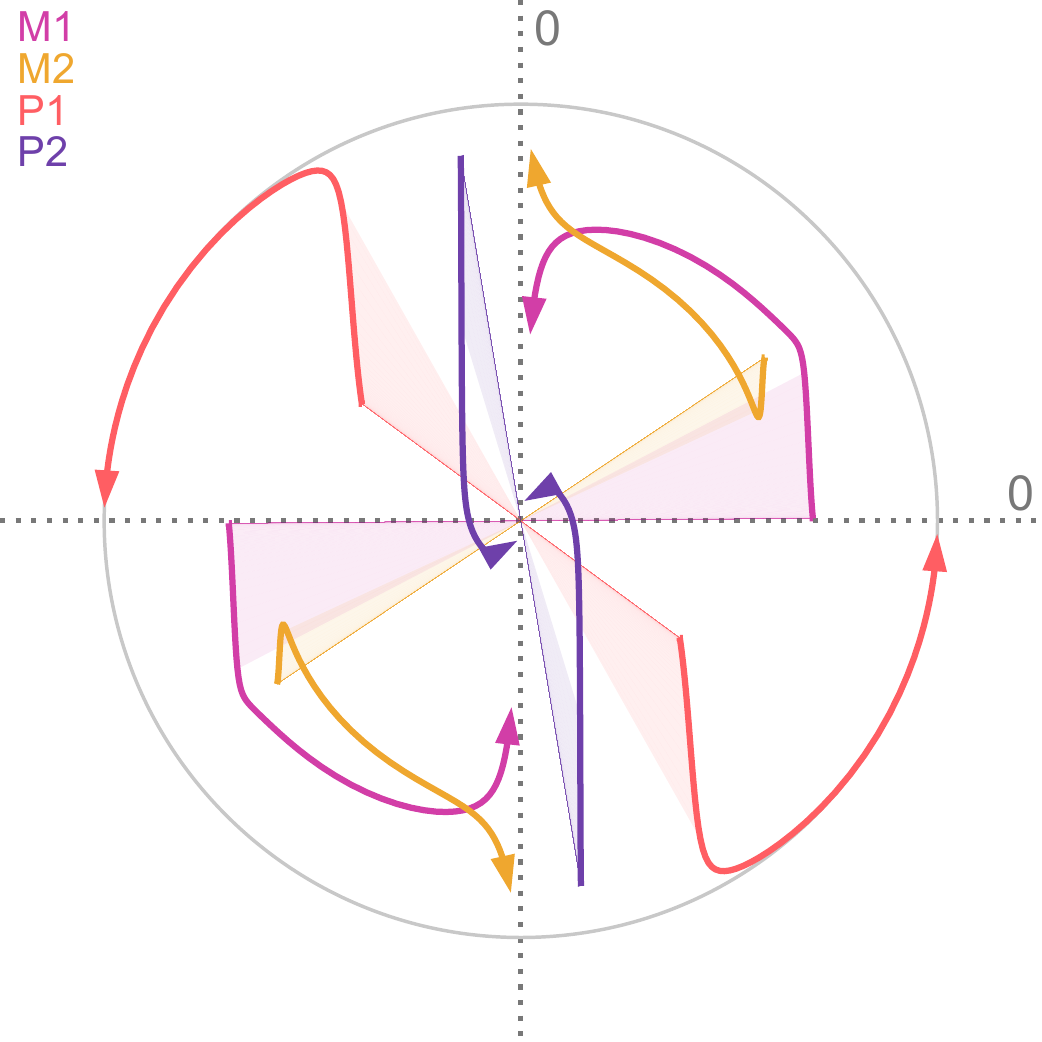}}}\,
  \subfloat[Dataset]{\plotframe{\includegraphics[width=0.483\linewidth]{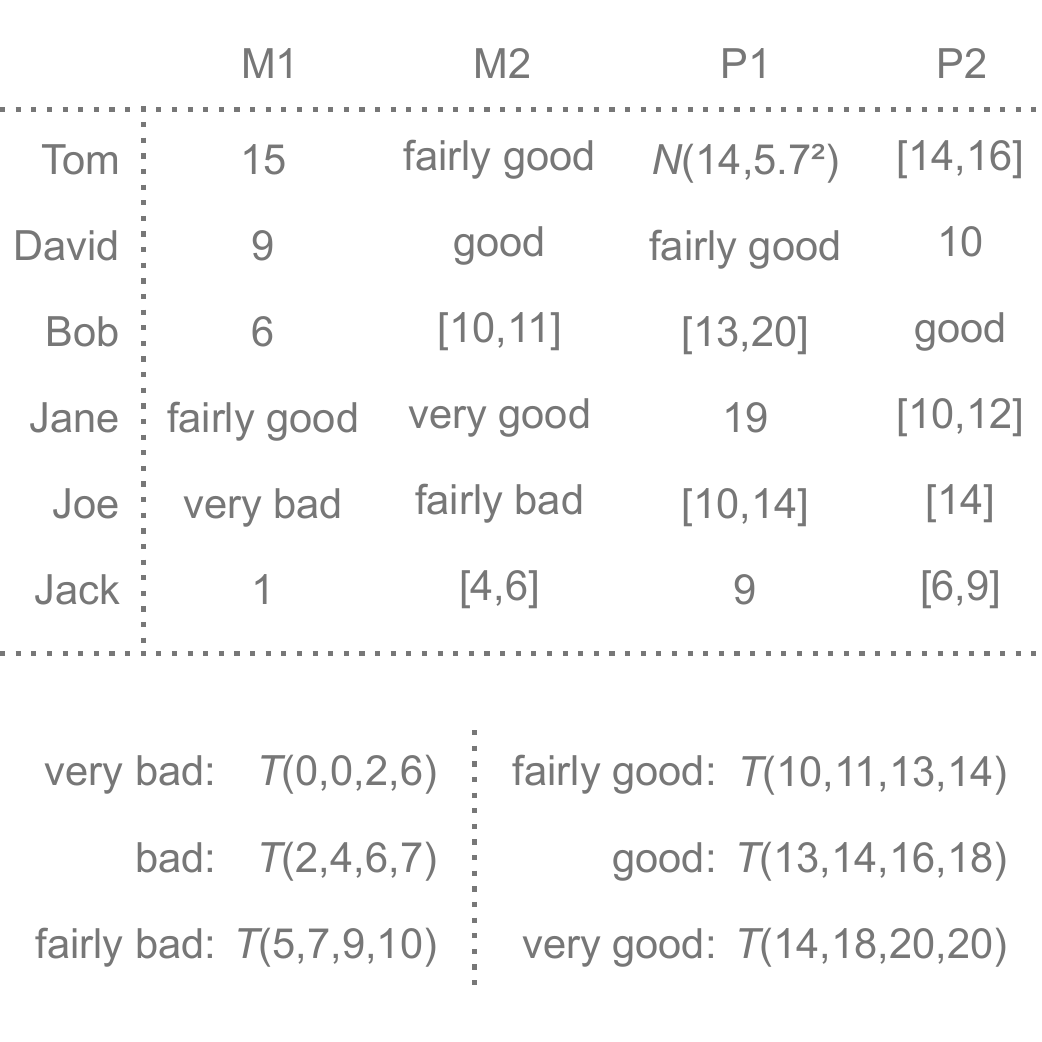}}}
  \caption{%
    Importance of respecting the uncertainty in the input data.
    (a)~Traditional PCA on the mean values of the student grades.
    (b) The projection found by our method shows the uncertainty in the data more faithfully.
    (c) The corresponding factor traces allow us to analyze the role of the original axes.
    In this case, \emph{P1} approaches unit length, which means that its information is present even after projection.
    (d) The dataset in tabular form, providing the trapezoidal distributions for the linguistic labels.
    A trapezoidal distribution \textit{T(a,b,c,d)} is defined by its bounds \textit{a}, \textit{d} and its discontinuities \textit{b}, \textit{c}.
  }
  \label{fig:student}
\end{figure}

\section{Examples}

Our method can handle various types of data uncertainty.
Following the classification of Skeels et al.~\cite{Skeels2009}, we will take a look at examples from the measurement precision level and the completeness level.
Measurement precision can play a substantial role in the analysis of datasets, especially for qualitative studies and experiments, where it is hard to assign certain values to responses.
One way to deal with this uncertainty is to assign fuzzy numbers or even explicitly encoded probability distributions to each of the data points, as we will show in Section~\ref{sec:students}.
Furthermore, we will look at different types of aggregations as sources for uncertainty on the completeness level.
Apart from these examples, we see potential use cases for our method in visualizing preprocessed data for real-time analysis, or data that has been aggregated to protect the privacy of individuals, such as medical data.
Regarding aggregation, Section~\ref{sec:discussion} gives more details about the computational complexity of our approach.
Please note that in the following examples, we use different representations for the distributions to highlight the projections found by our method.

\subsection{Student Grades}\label{sec:students}

Our uncertainty-aware PCA method can be used to perform dimensionality reduction on data with explicitly encoded uncertainty.
Amongst others, such data can be found in the domain of fuzzy systems.
As an example, we adopt the synthetic student grade dataset established by Denoeux and Masson~\cite{Denoeux2004}.
It consists of four test results (\emph{M1}, \emph{M2}, \emph{P1}, \emph{P2}) for each of six students.
The possible marks for the tests range from 0 to 20, and the dataset is highly heterogenous:
grades can be represented either as real numbers, such as $15$, without any uncertainty, or as intervals, such as $[10,12]$.
Furthermore, many grades are given by qualitative statements like \emph{fairly good} or \emph{bad}.
Both intervals and linguistic labels contain uncertainty, modeled using uniform distributions and trapezoidal distributions, respectively.
The original paper also contains one \emph{unknown} value.
We model the missing value using a normal distribution $\Normal(14,5.7^2)$, which we extract from prior information:
the mean is similar to previous test results, and the variance represents realistic deviations in both directions from this mean.
Figure~\ref{fig:student} shows the PCA on this dataset.
It is important to note that PCA performed solely on the means of the input, as shown in Figure~\ref{fig:student}a, fails to capture important uncertainty information in the data.
Our method (Figure~\ref{fig:student}b) appropriately depicts the uncertainty that is present in \emph{P1} of \emph{Tom} and \emph{Bob}.
This draws a very different picture from the result of regular PCA because the topology changes:
it is quite possible that \emph{Tom} performed similar to \emph{Jane}---a fact that is not readily visible from Figure~\ref{fig:student}a.
The importance of \emph{P1} on the resulting projection can also be seen in the factor trace (Figure~\ref{fig:student}c) for this dataset:
with an increasing amount of uncertainty factored into our method, the trace of \emph{P1} moves toward the outside of the unit circle.
The interpretation for this is that most of the information of this axis is preserved after projection.

\subsection{Iris Dataset}\label{sec:iris}

\begin{figure}[bt]
  \centering
  \subfloat[PCA on distributions]{\plotframe{\includegraphics[width=0.483\linewidth]{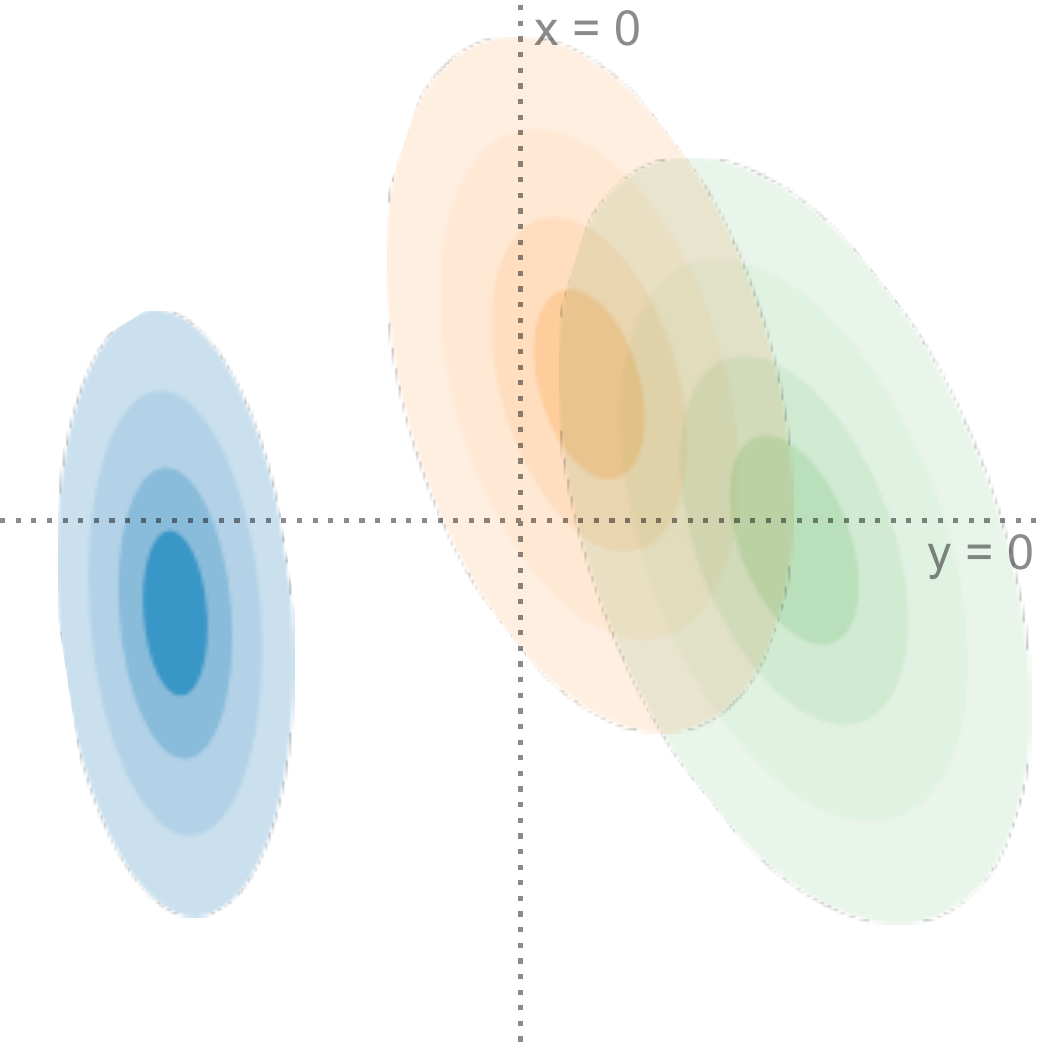}}}\,
  \subfloat[PCA on orginal points]{\plotframe{\includegraphics[width=0.483\linewidth]{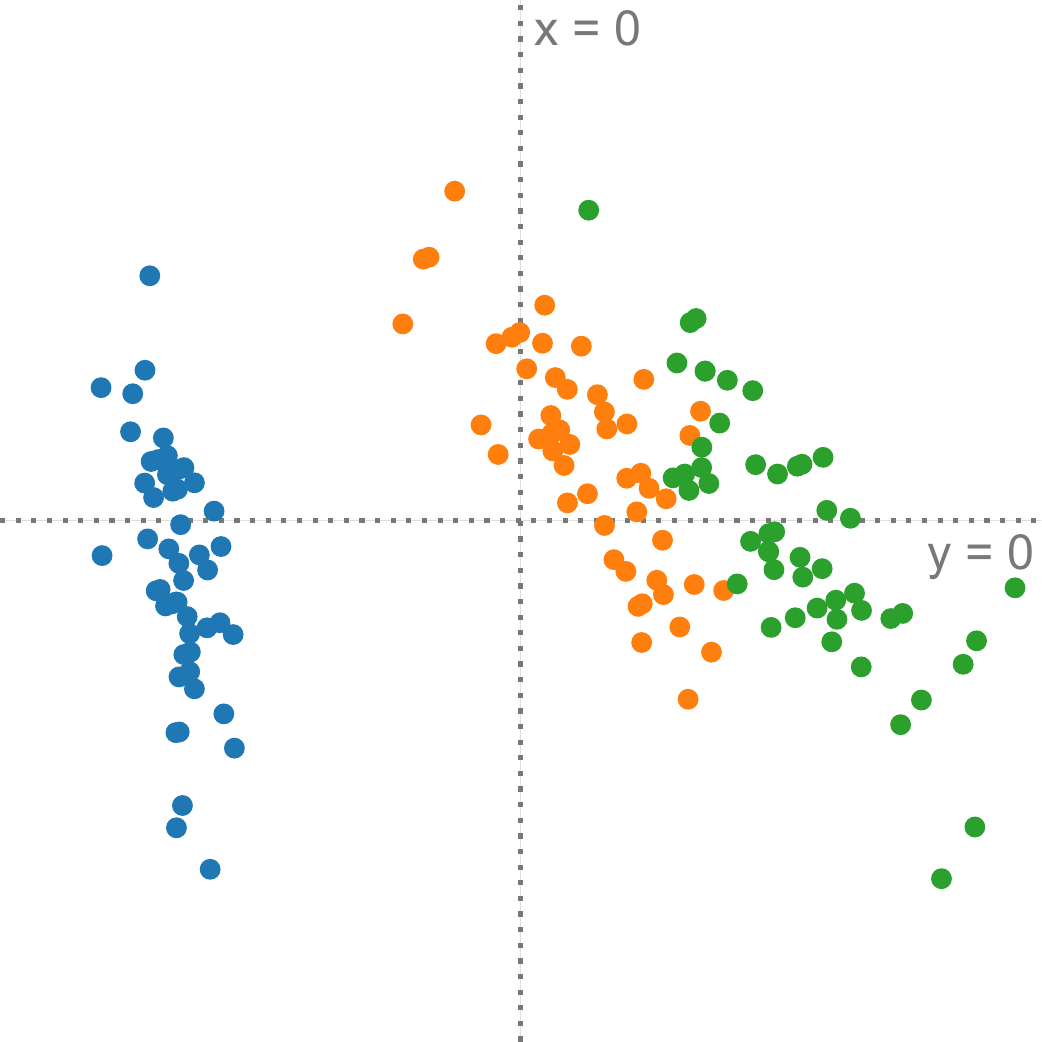}}}
  \caption{%
    Comparison between (a) our approach and (b) performing PCA on the original set of points of the Iris datasets.
    In (a) the aggregation into clusters has been performed before the projection, while in (b) the aggregation into the different clusters is performed visually through color.
  }
  \label{fig:iris}
\end{figure}

The Iris dataset\footnote{\url{https://archive.ics.uci.edu/ml/datasets/iris}} has widely been used to study projection and machine learning algorithms.
It is four-dimensional and consists of 150 specimen of the Iris plants.
Additionally, each instance can be attributed to one of three classes, and the instances are distributed equally among the classes.
The clusters of the Iris dataset can be well described using multivariate normal distributions.
We aggregate the data into three distributions, by their class label, on which we then perform uncertainty-aware PCA.
The result of this can be seen in Figure~\ref{fig:iris}a.
For comparison, we also perform conventional PCA and color each point according to its class label---the results are shown in Figure~\ref{fig:iris}b.
Both projections are almost identical.
This shows that our method can find projections with only a fraction of the original 4D data:
three multivariate normal distributions instead of 150 points.

This example also illustrates two different ways to visualize data that has additional labels.
To convey the class information, we need to support the visual aggregation of each cluster.
When using conventional projection methods, this aggregation is usually performed in the image space.
Figure~\ref{fig:iris}b, for example, uses color.
Another technique that is commonly used for aggregation in the image space is kernel density estimation.
For clusters that roughly follow a normal distribution, our method provides a different approach:
it allows aggregation in the feature space, where all the information is still present, and subsequent projection of the aggregated information.
Subsequently, no further aggregation has to be performed in the image space.
In Section~\ref{sec:sampling}, we provide a more detailed comparison to sampling-based strategies.

Figure~\ref{fig:eigenvalues}a shows the factor traces for the Iris dataset.
Here, we can see that \emph{petal width} moves closest to the center of our visualization.
This means that the dimensionality reduction, projects along this axis, especially for $s \rightarrow \infty$.
Furthermore, \emph{sepal width} and \emph{petal length} have almost no shaded area.
Because we use the shaded area to encode and highlight the interval $s \in [0,1)$, this illustrates that the projection of these two axes remains almost the same while interpolating between regular PCA and our method.

\subsection{Anuran Calls Dataset}\label{sec:anuran}

\begin{figure}[bt]
  \centering
  \subfloat[Original points colored by family]{\plotframe{\includegraphics[width=0.483\linewidth]{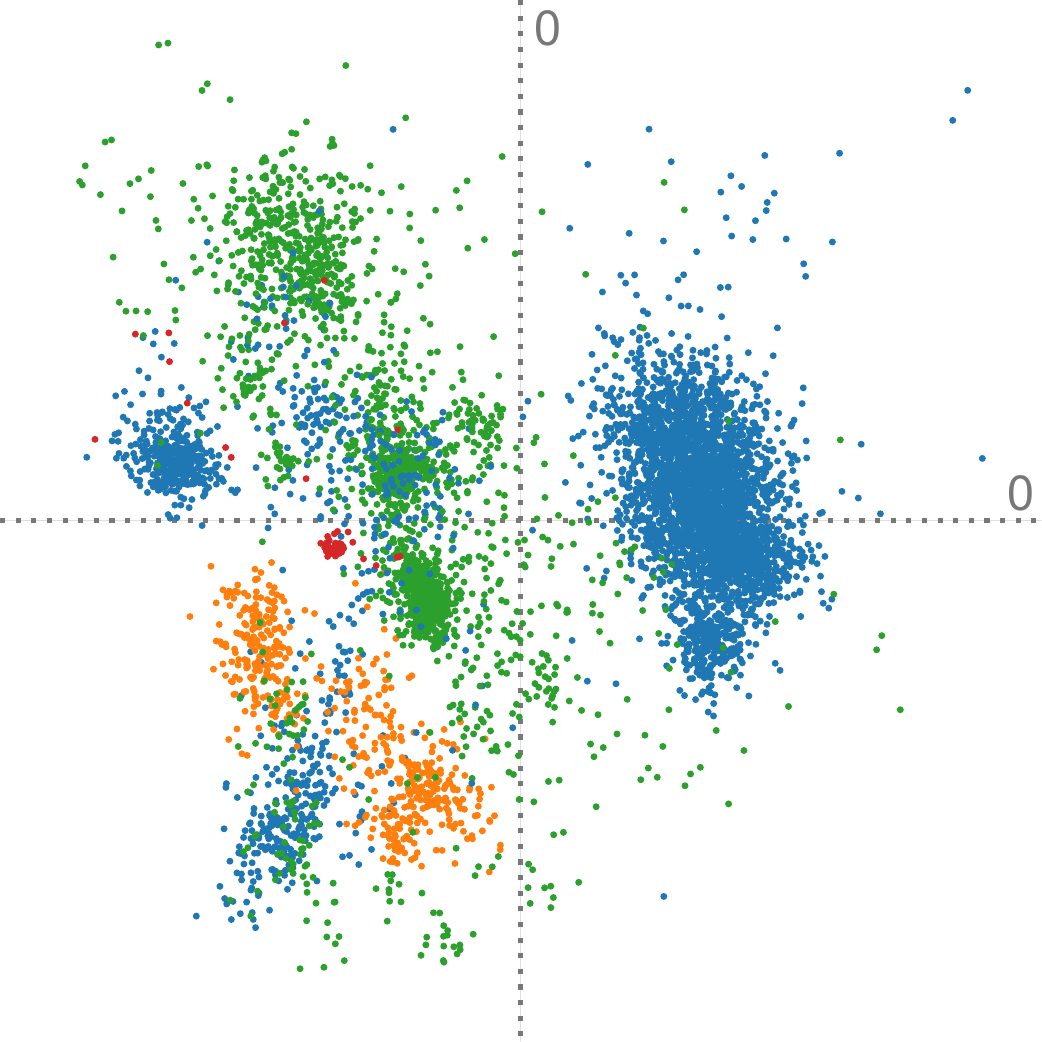}}}\,
  \subfloat[Distributions clustered by family]{\plotframe{\includegraphics[width=0.483\linewidth]{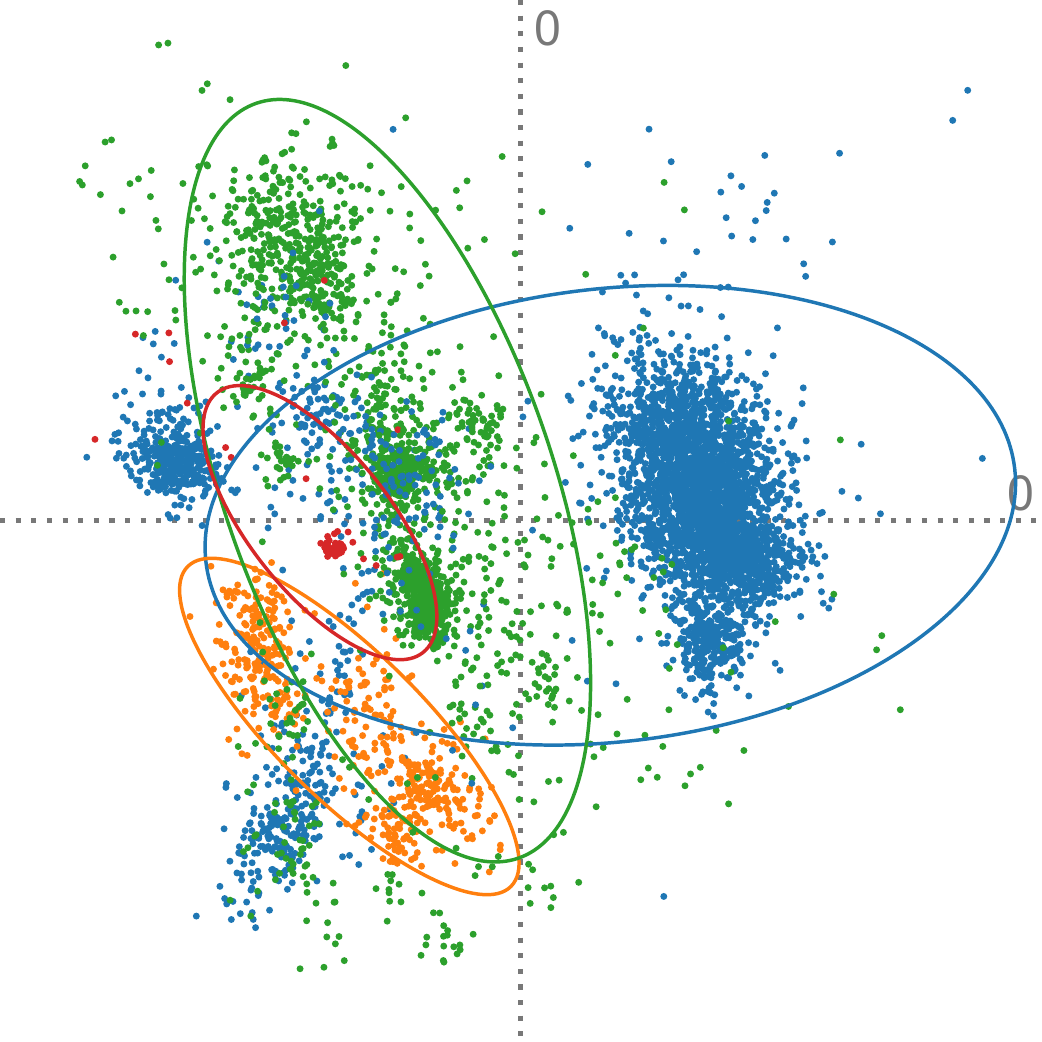}}}\\
  \subfloat[Original points colored by genus]{\plotframe{\includegraphics[width=0.483\linewidth]{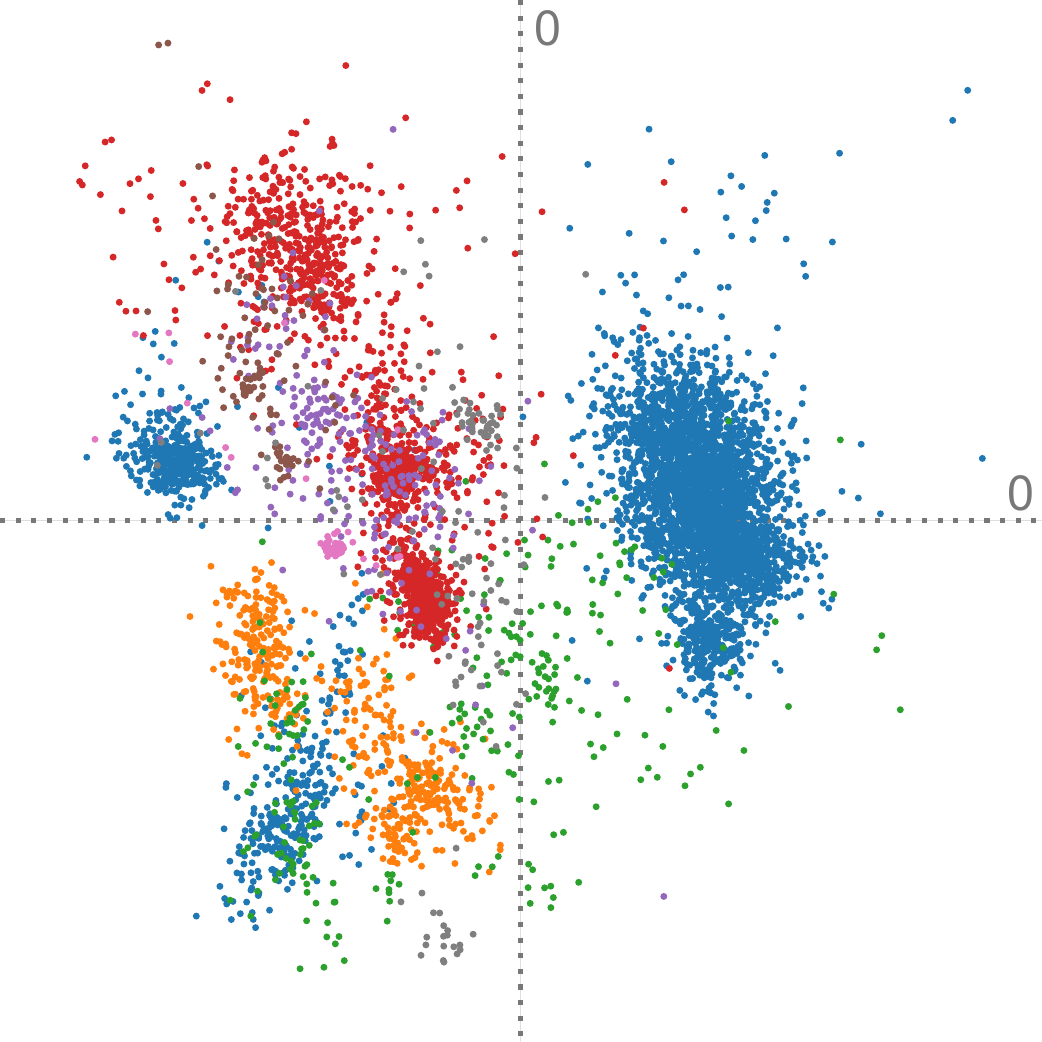}}}\,
  \subfloat[Distributions clustered by genus]{\plotframe{\includegraphics[width=0.483\linewidth]{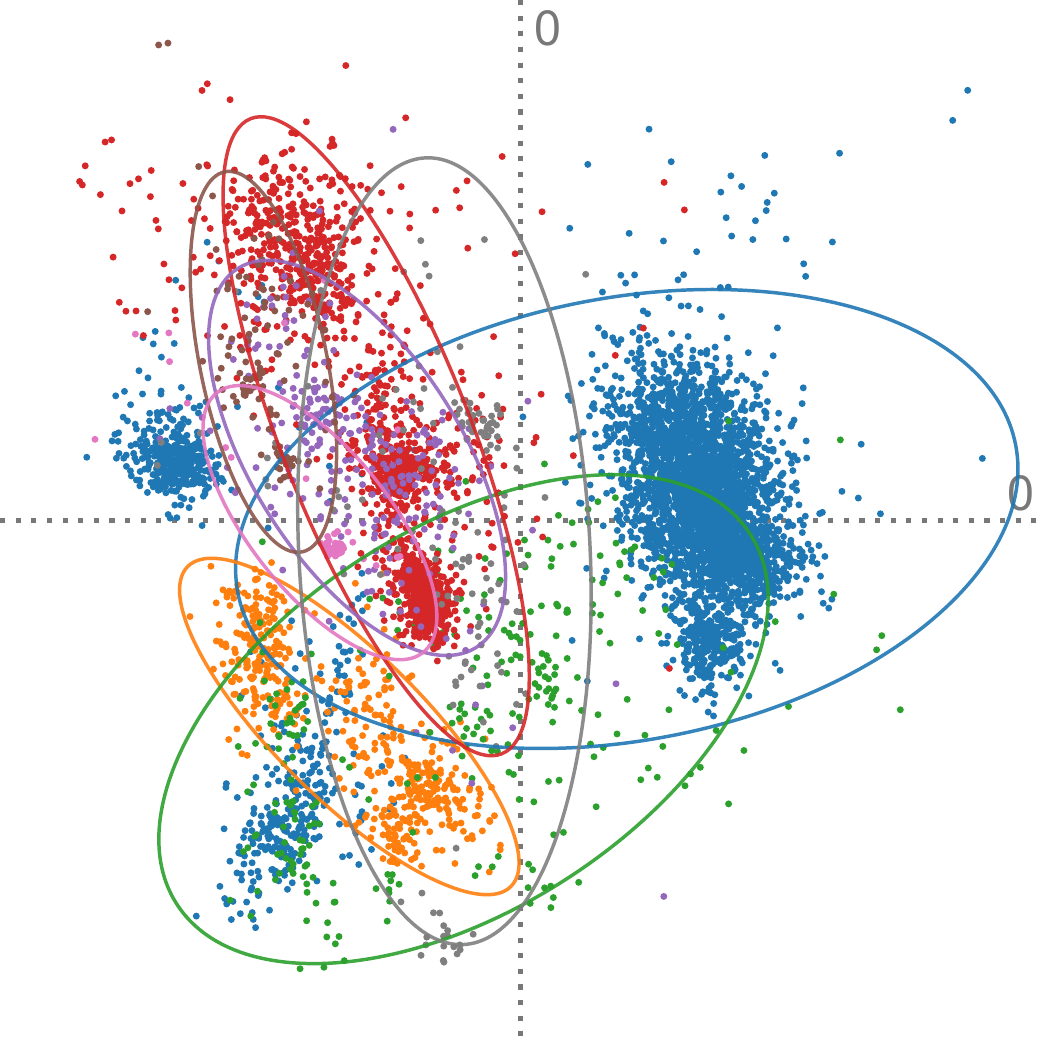}}}
  \caption{%
    Comparison of projections resulting from conventional PCA and our method.
    Projections of the extracted covariances are shown as ellipses.
    Clustering by family leads to four clusters, while clustering by genus results in eight clusters.
    Although the clusters have a large variance in the number of instances (a), our weighted approach matches the projection of the original dataset well.
    The projection also remains stable for clustering by a different class label, here, by genus (b).
    Overall our method (d) performs well, even though not all clusters in the original dataset follow a normal distribution (c).
  }
  \label{fig:anuran}
\end{figure}

The Anuran Calls dataset\footnote{\url{https://archive.ics.uci.edu/ml/datasets/Anuran+Calls+(MFCCs)}} contains acoustic sound features extracted from frog recordings.
In total, there are 7195 instances of such calls, and they are grouped by family, genus, and species labels.
Again, we perform aggregation of the instances, in this case, by looking at the family class label.
However, the interesting aspect of this dataset is that, in contrast to the Iris dataset (Section~\ref{sec:iris}), there is a different amount of instances per class.
There are calls from four different frog families in this dataset---the numbers of instances per class are 4420, 2165, 542, and 68.
These families can further be subgrouped by genus, yielding eight distinct clusters.
Furthermore, it is important to note that many groups do not follow a normal distribution and exhibit varying modality, as can been seen in Figure~\ref{fig:anuran}.

So far, we have assumed that all aggregated distributions represent the same amount of instances.
This can lead to overemphasized clusters if their original sample count is small.
Concerning this dataset, this would mean that the family with 4420 instances would receive the same amount of weight as the family with 68 instances.
To achieve a better fit to the actual data that these distributions stand for, we can adapt our method to take class weights into account by slightly modifying Equation~\ref{eq:final}.
In particular, it suffices to use the weighted average to evaluate $\EvCov{\cdot}$.
The computation of the sample mean needs to be adjusted accordingly.

Figure~\ref{fig:anuran} shows the comparison of our method, adapted to handle cluster weights, to regular PCA on the original set of points.
For Figure~\ref{fig:anuran}ab, the data is clustered by \emph{family}, yielding four distributions.
Figure~\ref{fig:anuran}bc was aggregated by \emph{genus}, which results in eight distinct discrete probability distributions.
For the projections that were created using our method, we show the covariances that were extracted from each of the different clusters.
This demonstrates that even if the clusters do not follow a simple distribution, such as the blue cluster in Figure~\ref{fig:anuran}b, our technique is still able to reconstruct the original PCA.
The projections that are found for the point data and the aggregated data are visually the same.
Assigning weights to each cluster according to the amount of data that it represents is an obvious application of this extension to our method.
However, we can also imagine that this technique can be used in a more exploratory setting, for example, by investigating the effect of one cluster on the resulting principal components.

\section{Comparison To Sampling}\label{sec:sampling}

In this section, we provide a comparison of our method with another strategy that could be used to construct the covariance matrix for uncertain data: sampling.
Instead of directly computing $\Cov{\ObSet}{\ObSet}$ on the distributions, we can draw samples from each of them.
If we concatenate the resulting set of points, we can use the conventional way for computing the covariance matrix as specified by Equation~\ref{eq:cov}.

To compare the resulting covariance matrices, we need a suitable distance metric.
We choose the Hellinger distance.
It is commonly used to compare the results of linear models~\cite{Torgersen1991}.
This distance metric is typically used to compare two multivariate normal distributions $p$ and $q$.
It is based on the Bhattacharyya coefficient, which can be used to describe the overlap between $p$ and $q$:
$$
BC(p,q) = \int\sqrt{p(x)q(x)}\:dx
$$
The Mahalanobis distance is a special case of the Bhattacharyya distance ($-\Ln(BC(p,q)$) for distributions that share the same covariance.
Using the definition of the Bhattacharyya coefficient, the Hellinger distance is defined as
$$
H(p,q) = \sqrt{1 - BC(p,q)}
$$
To apply this distance metric to the problem of comparing the results from principal component analysis, it is important to note that PCA is completly defined by its sample mean and overall covariance matrix.
Together, we interpret these two artifacts as a multivariate normal distribution.
The resulting distribution can then be compared using the Hellinger distance.
In contrast to a description based on eigenvalues and eigenvectors, our method is invariant against flipping and no further preprocessing has to be performed.

For our experiment, we applied PCA to a synthetic dataset with 10 distributions $\ObSet_{Syn} = \{\Ob_1, \dots, \Ob_{10}\}$, each following a normal distribution $\Ob_i \Follows \Normal(\Rvec{\DistMean_i}, \DistCov_i)$.
All the means $\DistMean_i$ are drawn from another overarching multivariate normal distribution:
$$
\Rvec{\DistMean_i} \Follows \Normal(\Zero, \Sigma)
$$
The covariance of each of the distributions $\DistCov_{i}$ is constant across the dataset.
It is created by reversing the elements of $\Sigma$.
Because of this, all covariances also share the same determinant.

\begin{figure}[t]
  \centering
  \plotframe{\includegraphics[width=\linewidth]{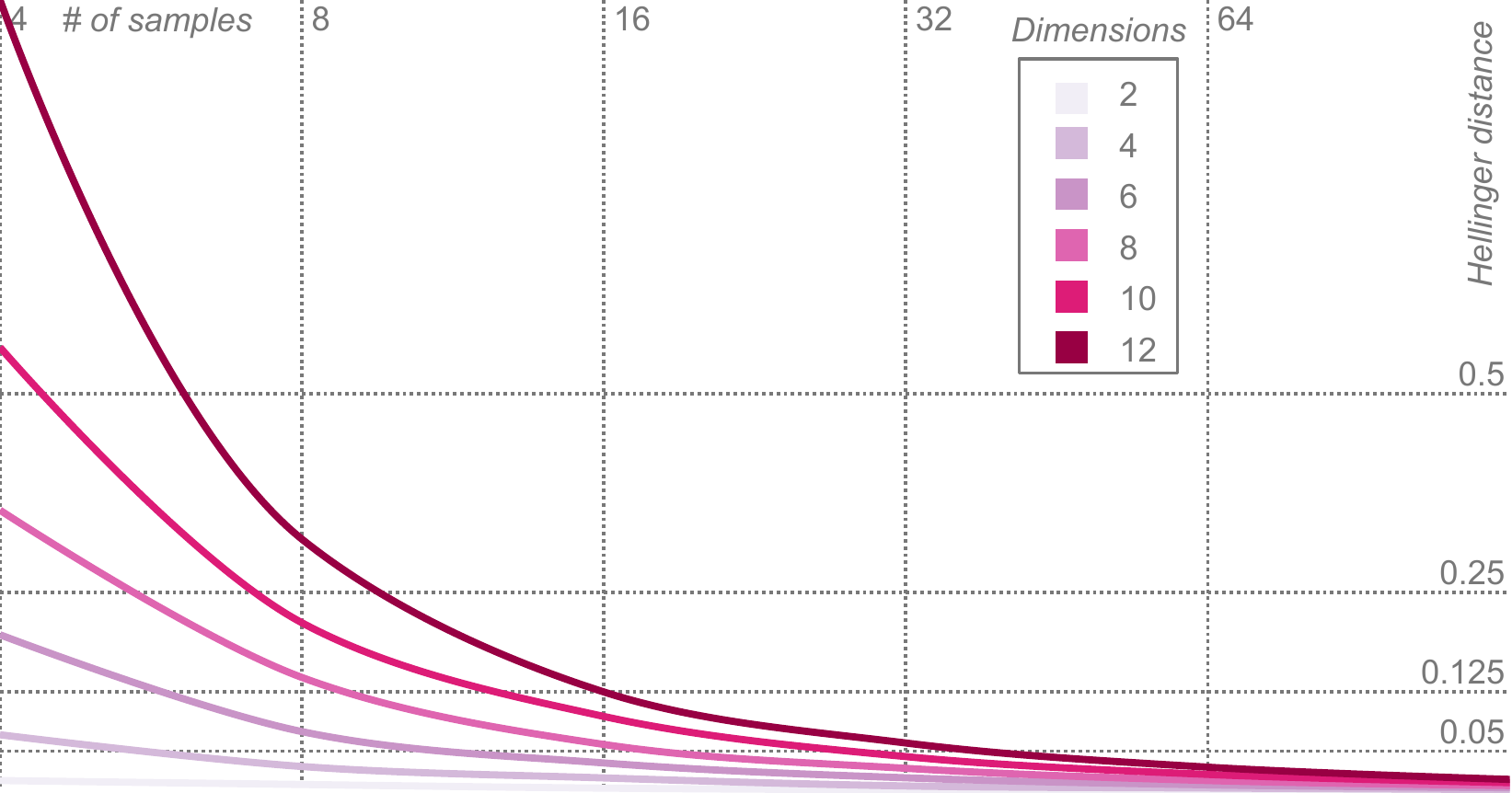}}
  \caption{%
    Multiple comparison of our method to a sampling-based approach using the Hellinger distance for input data with two to 12 dimensions.
    The \textit{x} axis shows the increasing number of samples that were used for the sampling strategy, while the \textit{y} axis shows the distance to the result from our method.
    The required number of samples for a good result grows with the number of dimensions.
  }
  \label{fig:sampling-comparison}
\end{figure}

Figure~\ref{fig:sampling-comparison} shows the results of our experiment.
For each data point we performed 40 runs and chose the median outcome.
We can draw several conclusions form our experiment.
First, it shows that the sampling approach converges to our method with an increasing number of samples.
This indicates that our method is a valid way to compute PCA on probability distributions.
Second, it shows that our method scales far better than the sampling-based approach with a growing number of dimensions.
We expect the curse-of-dimensionality to be the reason for this.

\section{Discussion}\label{sec:discussion}
In the following, we discuss the uncertainty-aware extension of PCA that we introduced, demonstrated, and assessed above from different perspectives.
To begin with, we compare its computational complexity to traditional PCA.
We follow up with more details on its application to interactive visualization, especially concerning scalability.
Finally, we discuss the general limitations of PCA and how these carry over to our method.

\subsection{Computational Complexity}
In general, our method has the same computational complexity as regular PCA.
For a dataset with $N$ samples and $D$ features, regular PCA has a computational complexity of $\BigO{ND^2}$ for the computation of the covariance matrix.
Retrieving the eigenvalues and eigenvectors has a complexity of $\BigO{D^3}$.

With our method, samples $N$ are $D$-dimensional probability distributions instead of points.
In many cases, the probability density function of a random vector $\Ob_n$ is known analytically, and $\Ev{\Ob_n}$ as well as $\Cov{\Ob_n}{\Ob_n}$ can be looked up in constant time $\BigO{1}$.
Our adapted computation of the global covariance matrix can be performed in $\BigO{2 \cdot ND^2}$ since we additionally need to compute the average covariance matrix over all $N$ distributions.
Asymptotically, however, the constant factor $2$ can be neglected.
This results in a complexity of $\BigO{ND^2}$ for determining the covariance matrix.

We share the extraction of the eigenvalues and eigenvectors with regular PCA.
As mentioned above, this can be performed in $\BigO{D^3}$.
Thus, our technique is of similar complexity as standard PCA.
Please note that in this analysis, we consider the aggregation of clusters as a preprocessing step (more details in the next section).
Its complexity would add to the total complexity, but is not considered here.
In the following section, we provide details on why preparing clusters is of special importance for the application of our technique to data visualization.

\subsection{Interactive Visualization and Scalability}
Big data is gaining relevance, and the amount of data that can be acquired and stored grows rapidly.
For example, the Large Hadron Collider (LHC) at CERN exceeded 200 Petabytes of collected sensor data already in 2017~\cite{Cern2017}.
At the same time, it often is critical to visualize such data for exploration, analysis, and knowledge generation~\cite{Sacha2014}.
Processing latencies are of significant concern for interactive visualization regarding big data.
We tackle this problem by separating the computationally complex task of data aggregation from the projection and visualization tasks.
Since our method is aware of the shape of the distributions, we can approximate the projection of clustered datasets by the projection of their respective distributions.
For a large number of samples $N$ in a $D$-dimensional feature space, this aggregation step is computationally costly since the covariance matrices have to be computed in $O(N D^{2})$.
The advantage of our method is that the aggregation can be done instantly during data acquisition and, in case memory demands are of concern, there is even no need to store raw data persistently~\cite{Wang2017}. In some fields, it is already common practice to aggregate data as a preprocessing step, for example, the \emph{in-situ} analysis in large-data visualization~\cite{Dutta2017}.
Using our method, the characteristics of the data are preserved during the complete pipeline, and its influence on the projection can still be taken into account during the analysis process.
Please note that when a cluster of multiple data points is aggregated by abstracting it as a normal distribution, the estimation of the covariance matrix is an inevitable step.
To do so, the number of data points needs to be sufficient concerning the number of dimensions, and there must not be problems with (local) outliers~\cite{Shevlyakov2011}.
Similarly, a small number of clusters can be a problem in high-dimensional space~\cite{Johnstone2009}.
By scaling the uncertainty of each cluster depending on the number of data points, it contains, our method compensates for differences in cluster sizes, as outlined in Section~\ref{sec:anuran}.
However, more research needs to be done in the direction of assessing whether the additional information provided by each clusters' weight and error covariance matrix can fully counter this problem.

\subsection{Limitations of PCA}\label{sec:limitations}

In practice, PCA is applied to all kinds of datasets, where it is commonly used as a tool for exploratory analysis.
Conceptually, our approach yields a projection operator that is more aware of the uncertainty in the data.
Just as with other linear methods, important information that is present in the non-principal components gets discarded due to the orthographic projection, which can guide the analysis into the wrong direction.
Our method inherits this limitation.
For regular PCA, methods have been developed to mitigate these effects---we provide an overview in Section~\ref{sec:related-work}.
For one, this is because one of the terms of our method essentially performs PCA on the expected values of each of the distributions, as described in Section~\ref{sec:uncertain-covariances}.
With regard to the uncertainty in the data, a second limiting factor can arise:
if the fraction of the covariance introduced by the uncertainty in the data is small in comparison to the covariance introduced by the expected values, and if the uncertainty happens to be orthogonal to the projection, it can also remain covert in the final representation.
Future research may investigate how non-linear methods, which could alleviate this problem, can be generalized to probability distributions too.

Several other factors pose challenges to finding the correct principal components.
The presence of outliers in the data can strongly influence the resulting projection.
This stems from the quadratic term in the computation of the covariance matrix. 
When outliers are of concern, forms of \emph{Robust PCA} (see Section~\ref{sec:related-work}), which rely on solving optimization problems, can be applied.
It remains to be seen how similar approaches can be adapted to uncertainty-aware PCA.
Although PCA was originally developed for real-valued data, it is often also used on datasets where some of the axes represent ordinal, and sometimes even categorical values.
Naturally, these axes can contain uncertainty information as well.
Furthermore, as of now, we do not explicitly model missing values.
In the context of regular PCA, several techniques have been developed to deal with this---Dray and Josse~\cite{Dray2014} provide a summary of approaches that can be applied in this case.
One straightforward way to handle these inputs in our framework nonetheless is imputation, as we have done in the student grade example provided in Section~\ref{sec:students}.
With our method, these imputed values can even take the form of more complex distributions, which is why we see this as a practical workaround.


\section{Conclusion}
In this paper, we have presented a technique for performing principal component analysis on probability distributions.
Unlike previous work, which mainly was concerned with non-correlated error models, our method works on arbitrary distributions.
We achieve this by incorporating first and second moments of the uncertain input data into the calculation of the global covariance matrix.
Our formulation of the global covariance matrix offers the potential for various extensions to traditional PCA.
Particularly, in this paper, we have shown the application to aggregated datasets (Section~\ref{sec:iris} and Section~\ref{sec:anuran}) and datasets with explicitly encoded errors (Section~\ref{sec:students}).

Principal component analysis, and linear dimensionality reduction techniques in general, have the advantage over non-linear methods that the projections remain interpretable.
The principal components found by PCA are linear combinations of the axes from the original data space.
With our technique, scaling the influence of the covariances of each of the distributions allows us to perform sensitivity analysis concerning uncertainty.
The factor traces we propose are a visual method to assess how uncertainty in the data is reflected by the contributions of each original dimension to the principal components.
Further, our technique preserves the low computational complexity and clear algorithmic structure of traditional PCA.
This enables the assessment of uncertainty induced differences to the projection by sampling different parameters for scaling the uncertainty.
As a result, our technique constitutes a next step towards the earnest consideration of uncertainty in the analysis of high-dimensional data and forms the foundation for straightforward extensions in numerous directions.

\section*{Appendix}\label{sec:appendix}
We show that our method, provided by Equation~\ref{eq:final}, indeed yields a covariance matrix by looking at the different terms of this equation.
A matrix $\Mat{K}$ is positive semi-definite if $\BiLin{u}{K} \geq 0$, for every non-zero vector $\Vec{x}$.

\newtheorem{theorem}{Theorem}
\begin{theorem}
The outer product $\OuterS{\Vec{x}} \in \Real^{d \times d}$ of a vector $\Vec{x} \in \Real^d$ with itself always results in a symmetric, positive semi-definite matrix.
\end{theorem}

\begin{proof}
Let $\Vec{u} \in \Real^d$ be a nonzero vector. Using the definition of positive semi-definitness from above,
$$
\BiLin{u}{(\OuterS{\Vec{x}})} = (\Transpose{\Vec{x}}\Vec{u})^2 \geq 0.
$$
The symmetry follows from the definition of matrix multiplication.
\end{proof}

Our method differs from regular PCA in one term, which is defined in Equation~\ref{eq:expected-cov}:
In essence, this term computes the arithmetic mean of the covariance matrices $\Cov{\Ob_i}{\Ob_i}$ of each distribution $\Ob_i$.
A matrix is a covariance matrix if and only if it is \emph{symmetric} and \emph{positive semi-definite}.
By definition, $\Cov{\Ob_i}{\Ob_i}$ always satisfies this property.

\begin{theorem}
Let $\mathbf{K} = \{\Mat{K}_1, \dots, \Mat{K}_N\}, \Mat{K}_n \in \Real^{d \times d}$ be a set of covariance matrices, then the arithmetic mean of this set $\frac{1}{N}\sum_{n=1}^{N} \Mat{K}_n$ is a covariance matrix.
\end{theorem}

\begin{proof}
Let $\Vec{u} \in \Real^d$ be a nonzero vector and $\Mat{A}, \Mat{B} \in \Real^{d \times d}$ positive semi-definite matrices.
Both addition $\Mat{A} + \Mat{B}$, and multiplication with a scalar $k\Mat{A}, k \geq 0$ result in positive semi-definite matrices:
\begin{align}
  \BiLin{u}{(A+B)} &= \BiLin{u}{A} + \BiLin{u}{B} \nonumber \\
  \BiLin{u}{(kA)} &= k(\BiLin{u}{A}) \nonumber
\end{align}
Because of this and the properties of symmetric matrices, it follows that the arithmetic mean of $\mathbf{K}$ is a symmetric and positive semi-definite matrix and therefore also a covariance matrix.
\end{proof}

\acknowledgments{%
We are very grateful to Sven Kosub, who helped us flesh out the mathematical model presented in Section~\ref{sec:method}.
This work was funded by the Deutsche Forschungsgemeinschaft (DFG, German Research Foundation) – Project-ID 251654672 – TRR 161.
Additionally, this work has received funding from the European Union’s Horizon 2020 research and innovation programme under grant agreement No 825041.
}

\bibliographystyle{abbrv-doi-hyperref-narrow}
\bibliography{references}

\end{document}


\maketitle

\section{Notation}
Throughout the paper we use the following notation:

\vspace{1em}
\begin{tabular}{ c l }
  Symbol & Description \\
  \hline
  $k$        & Scalar \\
  $\Vec{x}$  & Vector \\
  $\Mat{A}$  & Matrix \\
  $\Rvec{x}$ & Random vector \\
\end{tabular}